\documentclass[runningheads]{llncs}
\AtBeginDocument{%
  }

\usepackage{graphicx}
\usepackage{xcolor}
\usepackage{xspace}
\usepackage{booktabs}
\usepackage{multirow}
\usepackage{subcaption}
\usepackage{amsmath,amssymb}
\usepackage{algorithm}
\usepackage{algpseudocode}
\usepackage{enumitem}
\usepackage{wrapfig}
\usepackage{array}
\usepackage[hidelinks]{hyperref}
\usepackage{cleveref}
\usepackage{lineno}

\usepackage[sectionbib,numbers,sort&compress]{natbib}

\newcommand{\speedup}[1]{{${1.13\times}$}}
\newcommand{\Stable}{\mathrm{Stable}}
\newcommand{\Unstable}{\mathrm{Unstable}}

\newcommand{\toolname}{\textsc{FastCert}\xspace}
\newcommand{\tightpara}[1]{\noindent{\textbf{#1}}}

\begin{document}
\title{Uncovering the Limits of Proof Sharing for Neural Networks}

\author{Kanak Das\inst{1} \and
Shubham Ugare\inst{2} \and
Bor-Yuh Evan Chang\inst{3}\thanks{Bor-Yuh Evan Chang holds concurrent appointments at the University of Colorado Boulder and as an Amazon Scholar.
This paper describes work performed at the University of Colorado Boulder and is not associated with Amazon.} \and
Sasa Misailovic\inst{2} \and
Gagandeep Singh\inst{2} \and
Manu Sridharan\inst{1}}
\authorrunning{K. Das et al.}

\institute{University of California, Riverside\\
\email{kdas006@ucr.edu, manu@cs.ucr.edu} \and
University of Illinois Urbana-Champaign\\
\email{sugare2@illinois.edu, misailo@illinois.edu, ggnds@illinois.edu} \and
University of Colorado Boulder \& Amazon\\
\email{evan.chang@colorado.edu}}

\maketitle
\begin{abstract}
Robustness verification of neural networks is increasingly important, due to their use in many critical domains.  In certain scenarios, proof sharing has been shown to accelerate incomplete verification techniques by reusing intermediate-layer abstract states, or \emph{templates}, across queries.  However, questions remain as to the robustness of template-based acceleration across varying network architectures, properties, datasets, and training methods.  In this work, we perform a systematic study of the effectiveness of template-based acceleration and its limits.  Our study shows that template subsumption rates can vary widely across scenarios.  We present a novel metric of \emph{jointly stable neurons} to explain this variation, showing that in some cases template-based techniques are very unlikely to provide any speedup.  Then, we present \toolname{}, a novel technique for \emph{automatically} distributing templates across neural network layers to increase performance impact, eschewing templates entirely if they are unlikely to produce a speedup. Across a large set of covering-design based $L_0$-verification tasks, \toolname{} achieved an average speedup of \speedup{} over an extant template-based reuse technique.
\end{abstract}
\section{Introduction}
Neural networks have emerged as the dominant paradigm for image classification, powering applications from autonomous driving to medical diagnosis. 
Despite their success, it is established that carefully crafted perturbations known as adversarial examples can cause networks to 
misclassify inputs, while remaining imperceptible to human observers~\cite{SzegedyZSBEGF13,GoodfellowSS14,Carlini017}.

To mitigate these risks, there has been significant interest in creating certifiably robust neural networks, resistant to adversarial threats~\cite{madry:17}. 
These methods expose networks to adversarial perturbations during training, often leveraging formal 
verification techniques to guide and refine the optimization process~\cite{wong2017provable,mirman2018differentiable}. Then, formal verifiers~\cite{wu2024marabou20versatileformal,li2023sok} are used post-training 
to guarantee the network's robustness to specified perturbations, such as sound but incomplete verifiers based on abstract interpretation~\cite{singh2018fast,singh2019abstract,wong2017provable}.

Recent work has shown that intermediate abstractions computed by such incomplete verifiers can be generalized into templates~\cite{10.1145/3527319,10.1007/978-3-031-13185-1_7}, 
enabling proofs to be \textit{reused} and accelerating subsequent 
verification queries. Typically, a small set of templates are derived by verifying local $L_\infty$ properties~\cite{SzegedyZSBEGF13,GoodfellowSS14,madry:17}, each permitting perturbations within a related
region of the input space. These methods have been shown to be effective on related properties of the same input over the same or quantized networks, such as shifted patches or random $t$-pixel perturbations over an image.  But, the effectiveness of template-based acceleration has not been studied carefully in the context of other properties such as $L_0$ verification, a realistic threat model~\cite{kotyan2022adversarial} where there have been significant recent advances in verification practicality~\cite{Shapira23,shapira2024boostingfewpixelrobustnessverification}.  Further, no work has studied the limits of template-based acceleration across a variety of network architectures, properties, data sets, and training methods.

Our first key contribution is a systematic framework characterizing when template reuse is a principled accelerator for robustness verification and when the verifier should instead run without it.  For a template-based verifier to achieve speedups, a high percentage of verification queries must be subsumed by a small set of templates, but at the same time, each template must be precise enough to enable query verification. 
We devise a novel metric of \emph{jointly stable neurons} across a set of abstract states to explain the potential for template reuse given this tradeoff.  Through a limit study, we show that if the percentage of jointly stable neurons is low, extant template-based proof sharing techniques are unlikely to provide any verification speedup.  Our measurements across several networks and training methods show that the potential for template-based acceleration varies widely across scenarios.

Given the observed variance in template effectiveness, there exists a need to automatically determine how best to employ templates for a given verification task.
Our second key contribution is a novel technique \toolname{} to \emph{automatically} distribute templates across layers of a neural network to maximize performance impact, or eschew templates if they are unlikely to produce a speedup.  
\toolname{} performs profiling and template generation before verification, predicting template subsumption rates by sampling a very small set of queries. We implement \toolname{} and show that for a state-of-the-art $L_0$ verifier, it nearly always provides a greater speedup than previous proof-sharing techniques in cases where templates can be effective, while significantly mitigating the slowdown when templates are ineffective.

This paper makes the following contributions:

\begin{itemize}
  \item We perform a limit study of the effectiveness of template-based proof reuse, characterize its potential success using a novel metric of jointly stable neurons and showing the success rate varies widely.

  \item 
  We give a method to automatically determine whether, where, and how to apply template reuse for a given network and set of verification tasks.

  \item We implement our approach in a tool \toolname and demonstrate its practicality for accelerating covering-design-based $L_0$ verification, with an average speedup of \speedup{} over the state-of-the-art approach for applying templates (a 7 hour reduction in wall-clock time.).%
\end{itemize}

\section{Background}
\label{sec:background}

\tightpara{Neural Networks}
A neural network $N$ is a function $N: \mathbb{R}^{d_\text{in}} \to \mathbb{R}^{d_\text{out}}$, commonly built from a composition of
individual layers $N_L \circ N_{L-1} \circ \cdots \circ N_1$. We consider fully-connected feed-forward networks, where each layer 
$N_i(\mathbf{x}) = \text{ReLU}(\mathbf{A} \mathbf{x} + \mathbf{b})$.  The Rectified Linear Unit (ReLU) activation applies $\max(\cdot, 0)$ element-wise. For a classification task with $c$ classes, the network outputs $d_\text{out} := c$ scores,
assigning the class with the highest score as its prediction. For $k < L$, $N_{1:k}$ denotes the application of the first $k$ layers
and $N_{k+1:L}$ denote the final $L-k$ layers.

\tightpara{Local Robustness Verification}
Given a set of inputs and a postcondition $\psi$, neural network verification aims to prove that $\psi$ holds on the network output for all given inputs~\cite{albarghouthi2021introductionneuralnetworkverification}.
\emph{Local robustness verification} proves that $\psi$ holds for all network outputs corresponding to an input region
$\mathcal{I}(\mathbf{x}_0)$ formed around some input $\mathbf{x}_0$. Formally, local robustness
verification proves that $\forall \mathbf{z} \in \mathcal{I}(\mathbf{x}_0), N(\mathbf{z}) \models \psi$.
We write $\mathcal{I}(\mathbf{x}_0) \models \psi$ if the property holds.

\tightpara{Threat Models and Specifications}
The $L_0$ threat model captures \emph{sparse} perturbations: given an input $\mathbf{x}_0$,
the admissible region $\mathcal{I}_{L_0}(\mathbf{x}_0, k)$ is $\{\mathbf{z} \mid \| \mathbf{x}_0 - \mathbf{z} \|_0 \leq k \}$, allowing an adversary to arbitrarily modify up
to $k$ input features\cite{SuVS19,modas2019sparsefool,CroceASF022}. Unlike norm-bounded dense noise (e.g., small $L_\infty$ deviations) that yields a single convex region, the $L_0$ model induces a combinatorial family
of verification subproblems for one image, corresponding to the many choices of which features are altered.
In the patch threat model~\cite{ChiangNAZSG20,eykholt2018robustphysicalworldattacksdeep}, an adversary may arbitrarily perturb the pixels inside a single contiguous $w\times h$ region, 
leaving all other pixels unchanged; the input region $\mathcal{I}_{\text{patch}}(\mathbf{x}_0, w, h)$ allows arbitrary changes to any $w \times h$ patch within the image,
yielding multiple verification subproblems across patch locations. In geometric perturbations, the adversary applies a transform $T_{\delta}$ from a
prescribed family (e.g., bounded rotation/contrast/brightness), with parameters $\delta \in \Delta$; 
the input region $\mathcal{I}_{\text{geo}}(\mathbf{x}_0,\Delta)=\{T_{\delta}(\mathbf{x}_0)\mid \delta\in\Delta\}$ is typically non-convex due to
resampling/interpolation and is commonly verified by splitting $\Delta$ into finitely many $r$ cases~\cite{balunovic2019certifying}.

\tightpara{Robustness Under Adversarial Threats}
To defend against the adversarial threat, specialized types of neural network training is employed.
It pursues a dual objective: minimize classification loss on data while maximizing robustness so
predictions remain stable within the threat model. PGD~\cite{madry:17} training does this empirically by finding worst-case perturbations per example and
updating the model to classify them; IBP~\cite{mirman2018differentiable} training does it with sound guarantees by propagating bounds to certify that all inputs in a region
keep the label and optimizing the certified margin. Recent methods such as SABR~\cite{sabr} compute bounds on small, carefully selected subregions of the adversarial space
to reduce approximation error and improve both standard and certified accuracy, while TAPS~\cite{taps} combines IBP and PGD to optimize more precise but unsound worst-case loss approximations.
CURE~\cite{jiang2025universalcertifiedrobustnessmultinorm} targets multi-norm robustness by regularizing and shaping networks so bounds 
remain tight across multiple norms. For localized threats, certified patch defenses~\cite{ChiangNAZSG20}
extend guarantees from norm-bounded noise to contiguous regions by reasoning over all patch placements.

\tightpara{Verification via Abstract Interpretation}
Scalable neural network verifiers typically rely on \emph{abstract interpretation}~\cite{Cousot:77,singh2018fast,singh2019abstract,safetytrustabsint} to soundly overapproximate the set of reachable outputs.
The verifier first encodes the input region with a shape in some \emph{abstract domain} (e.g., boxes, zonotopes, or polyhedra) that encloses 
$\mathcal{I}(\mathbf{x}_0)$.
This shape is propagated layer by layer via \emph{abstract transformers} that overapproximate each layer's concrete operations.
After $k$ layers, the resulting \emph{abstract state} $S_k$ overapproximates the concrete states reachable at that layer.
At the output layer, $S_L$ bounds all possible network outputs; if it also satisfies the postcondition $\psi$, the property is certified to hold. This approach is sound but incomplete: the overapproximation may be too coarse to prove a valid $\psi$.

\section{Proof Templates}\label{motivating-example}
In this section, we describe template-based proof sharing for incomplete verification and present a simple performance model for when reuse is beneficial.

\subsection{Preliminaries}\label{sec:template-background}
Neural network verification via abstract interpretation has been optimized using \emph{proof templates}~\cite{10.1145/3527319,10.1007/978-3-031-13185-1_7}.
A template at an intermediate layer $k$ is an abstract state $T_k$ that has been certified to satisfy the postcondition $\psi$, i.e., $N_{k+1:L}(T_k) \models \psi$.
To use a template for a new verification query, the verifier propagates the query to layer $k$, obtaining abstract state $Q_k$.
If $Q_k$ is fully contained within the template, termed \emph{subsumption}, and denoted $Q_k \sqsubseteq T_k$, then $N_{k+1:L}(T_k) \models \psi$ implies $N_{k+1:L}(Q_k) \models \psi$, verifying the query without further propagation.

In practice, the box domain is employed for representing templates~\cite{10.1145/3527319,10.1007/978-3-031-13185-1_7}: at layer $k$, a box is a vector of intervals,
$S_k = \left\langle [l_j, u_j] \right\rangle_{j=1}^{n_k}$,
with each interval bounding the output of neuron $j$. Subsumption checking in the box domain is very efficient, making it desirable for templates.

\begin{figure}[t]
  \centering
  \includegraphics[width=\textwidth]{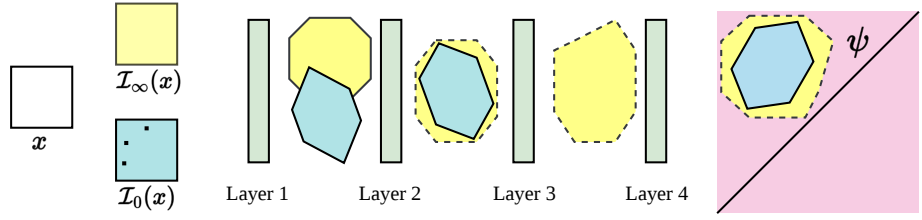}
\caption{Template reuse in a 4-layer network. From an input image $x$, we first verify an initial query under $\mathcal{I}_\infty(x)$, then 
build a layer-$k{=}2$ template by iteratively relaxing the layer-2 abstract state while it remains verifiable (represented by the yellow region with dotted boundary).
For a later query under $\mathcal{I}_0(x)$, the propagated layer-2 query state is subsumed by the template, so the property is proved immediately and propagation through layers 3 and 4 is skipped, saving tail cost.}

  \label{fig:motivating}
\end{figure}

\subsection{Template Construction and Use}
\label{template-generation}

In prior work~\cite{10.1007/978-3-031-13185-1_7,10.1145/3527319}, a verified query is leveraged to construct a proof template by generalizing its intermediate-layer abstract state.  In \cref{fig:motivating}, the abstract interpreter first verifies an $L_\infty$ query \(\mathcal{I}_\infty(x)\)---it computes an abstract state after each layer over-approximating the result of passing the input region through the previous layers.  
The query is considered verified if the final abstract state, computed after layer 4, satisfies the post-condition $\psi$. For \cref{fig:motivating}, as shown by the decision boundary after layer 4, the query is successfully verified.  

To construct a template, first, a set of layers is chosen at which to use templates.
At a selected layer $k$, the previously-computed abstract state is iteratively relaxed while preserving verifiability through the suffix \(N_{k+1:L}\),
yielding a more general state that can subsume subsequent queries whose layer-$k$ states are similar. Existing approaches commonly instantiate this procedure from
\(L_\infty\) specifications~\cite{10.1007/978-3-031-13185-1_7,10.1145/3527319}: they first use binary search to find the largest verifiable \(L_\infty\)-ball around the input, then relax the resulting layer-$k$ abstract
state to maximize generality subject to successful propagation and discharge on \(N_{k+1:L}\). The resulting relaxed state is stored as a template at layer $k$, which can be reused for potential subsumption of future queries.
Moreover, prior work shows that replacing a single \(L_\infty\)-ball over the input space with multiple masked \(L_\infty\)-balls over subregions yields a more diverse template
set and increases subsumption rates~\cite{10.1007/978-3-031-13185-1_7,10.1145/3527319}.

Using templates requires checking if the abstract shape computed for a subsequent query is subsumed by a template (see \cref{sec:template-background}).  For \cref{fig:motivating},
a template is placed after layer 2 (the yellow region with dotted boundary), and a subsequent query aims to verify an $L_0$ query \(\mathcal{I}_0(x)\).
Since the abstract state for this query is contained within the layer 2 template, the property is verified, without further propagation of the abstract state through the remaining layers.

\subsection{A Performance Model for Template-Based Verification}\label{sec:template-model}

The potential performance improvement provided by proof sharing with templates is impacted by various factors, and depending on the template subsumption rate, templates may even lead to a slowdown (which we observed in practice).  We will refer back to the model presented here when discussing the potential for template-based speedups for different networks (\cref{sec:proof-sharing-potential}) and presenting our technique for automatically optimizing template usage (\cref{sec:diagnostic}).

\label{sec:speedup-model}

Consider a network \(N\) with \(L\) layers and a fixed input \(x\) inducing \(r\) verification specifications. Fix a reuse layer \(k\) and a template set \(\mathcal{T}\) of size \(m\). Let \(t_{\mathrm{gen}}\) denote the one-time cost to construct \(\mathcal{T}\); \(t_{\mathrm{pref}}(k)\) the per-specification cost to propagate through \(N_{1:k}\); \(t_{\mathrm{match}}(m)\) the per-specification cost to test subsumption \(Q_k \sqsubseteq T\) for all \(T\in\mathcal{T}\); and \(t_{\mathrm{tail}}(k)\) the per-specification cost to complete verification over \(N_{k+1:L}\) when no match occurs. Let \(\rho_k\in[0,1]\) be the fraction of the $r$ specifications whose layer-$k$ abstract state is subsumed by at least one template in $\mathcal{T}$. The expected end-to-end verification time with template reuse is then,
\begin{equation}
t_{\mathrm{gen}} + r\Bigl( t_{\mathrm{pref}}(k) + t_{\mathrm{match}}(m) + (1-\rho_k)\,t_{\mathrm{tail}}(k) \Bigr)
\end{equation}

Under the assumption that the baseline verification exhibits approximately constant per-layer computational cost $\nu$, it follows that
$t_{\mathrm{pref}}(k)=k\nu$,
$t_{\mathrm{tail}}(k)=(L-k)\nu$,
$t_{\mathrm{gen}}=\lambda m L\nu$ for template creation time parameter $\lambda>0$\,
and $t_{\mathrm{match}}(m)=\eta m$ for template matching time parameter $\eta>0$. Speedup $S$, relative to baseline verification time $r L \nu$ is therefore,

\begin{equation}
\label{eq:speedup-constlayer}
S \;=\; \Bigl(1 + \lambda \tfrac{m}{r} + \tfrac{\eta m}{L\nu} - \rho_k \tfrac{L-k}{L}\Bigr)^{-1}
\end{equation}

Hence \(S>1\) is precisely when the \emph{saved tail cost} \(\rho_k (L-k)/L\) exceeds the \emph{amortized overheads} of template generation and lookup, 
\(\lambda m/r + \eta m/(L\nu)\).
Operationally, this favors (i) \emph{earlier viable layers} where \((L-k)/L\) is large, provided the queries yields a sufficiently high \(\rho_k\);
(ii) \emph{a small number of templates}
to keep lookup and generation costs small; and (iii) \emph{large number} of specifications, which amortize \(t_{\mathrm{gen}}\).
Using templates at very deep layers (small residual) or oversized template sets (large overhead) eliminate gains, in which case the template based approach might not be beneficial.

With this speedup model, prior work~\cite{10.1145/3527319,10.1007/978-3-031-13185-1_7} has shown that template reuse can yield significant performance gains
for patch verification, fixed-size $L_0$ verification, and geometric perturbations. In such scenarios, verification instances count $r$ is typically a few hundred and template set length $m$ is up to 4.  However, prior work did not perform a rigorous study of the impacts of neural network type, dataset, training method, and layer choice on the effectiveness of template-based proof sharing.

\section{The Limits of Template-Based Proof Sharing}\label{sec:proof-sharing-potential}

Here, we first present a model for understanding the limits of existing template-based proof sharing techniques to speed neural network verification, based on the concept of \emph{neuron stability}.  Then, we present a limit study of template-based proof reuse, showing its applicability varies widely 
depending on specific settings.

\subsection{Efficient Proof Sharing and Neuron Stability}\label{sec:neuron-stability}

For efficient proof sharing via templates, \cref{sec:speedup-model} shows that we need (1) a high template match rate, (2) low cost to compute templates, and (3) low cost to determine template subsumption for an abstract shape.  Further, there is a baseline correctness condition that (4) the abstract interpreter can verify the property for the template.  Here, we explore how neuron stability interacts with these conditions, specifically how many unstable neurons make it very unlikely all four conditions can be simultaneously satisfied.

Neural network ReLU activation layers (see \cref{sec:background}) act element-wise, suppressing negative pre-activations and preserving positive ones. ReLU therefore partitions neurons into active (positive output)
and inactive (zero output) sets, creating an activation pattern.
In abstract interpretation, when propagating an abstract state through the network, ReLU layers are the primary source of imprecision. For the box domain, given an abstract state for a neuron $j$ represented by the interval $[l_j, u_j]$,
the abstract interpreter must determine the effect of applying ReLU on the interval.  This requires a case analysis based on the containment of 0 in $[l_j, u_j]$, as 0 is the ReLU decision boundary.  Hence, the precision of the abstract interpretation hinges on whether the abstract state of the pre-activation value of a neuron is sufficient to determine the outcome of its ReLU activation.

A neuron is considered \emph{stable}~\cite{Botoeva:20,serra2021scalingexactneuralnetwork} if its pre-activation interval, $[l_j, u_j]$,
does \emph{not} cross the ReLU's decision boundary of 0. This occurs under two conditions:

\begin{itemize}
    \item The interval is entirely non-negative ($l_j \ge 0$), meaning the neuron is \emph{always active} and the ReLU function consistently acts as the identity.
    \item The interval is entirely non-positive ($u_j \le 0$), meaning the neuron is \emph{always inactive} and the ReLU behaves as the constant zero function.
\end{itemize}

In both cases, the activation behavior is fixed across all concrete behaviors represented by the abstract state, hence the term \emph{stability}.

A neuron is \emph{unstable} if its interval $[l_j, u_j]$ has a negative lower bound and a positive upper bound ($l_j < 0 < u_j$).
For such neurons, the abstract state is too coarse to resolve the sign of the pre-activation value.
Consequently, a sound abstract transformer for the ReLU function must conservatively account for both outcomes (an active or inactive neuron),
typically leading to precision loss.  For an abstract state $S_k$, we denote the sets of indices for stable and unstable neurons as $\Stable(S_k)$ and $\Unstable(S_k)$, respectively.

\paragraph{Induced Partial Orders}

The subsumption relation $\sqsubseteq$ on abstract states (\cref{sec:template-background}) has direct implications for neuron stability.
If a state $Q_k$ is subsumed by a template $T_k$, any unstable neuron in $Q_k$ must also be unstable in $T_k$. This is because the query's interval $[l_j^Q, u_j^Q]$
containing zero implies the necessarily wider template interval $[l_j^T, u_j^T]$ also contains zero. This yields the following inclusion relation unstable neuron sets:
\[
Q_k \sqsubseteq T_k \quad \implies \quad \Unstable(Q_k) \subseteq \Unstable(T_k)
\]
Hence, a template must accommodate all the instabilities present in any query it subsumes. Similarly, the sets of stable neurons are related by the inclusion $\Stable(T_k) \subseteq \Stable(Q_k)$.

\paragraph{Template Generality vs. Stability Variance}
The effectiveness of proof sharing hinges on a trade-off between template generality and the stability of individual queries.
For a small set of templates to be effective, each template must be general enough to subsume a large number of queries. But, for a single template $T_k$ capable of subsuming a set of queries $\{Q^i_k\}_{i=1}^m$, $\Unstable(T_k)$ must include the \emph{union} of the unstable neurons of the queries:
\begin{equation}
  \bigcup_i \Unstable(Q^i_k) \subseteq \Unstable(T_k)
  \label{eq:unstable-union-inclusion}
\end{equation}

Hence, a very general template must account for all the
instabilities present in the queries it subsumes.
This instability set can grow large if the queries exhibit \emph{diverse} instability patterns, with differing unstable neurons across the abstract states for individual queries.%
\footnote{In principle, a neuron that is stable in each individual query can still be forced unstable in a template if it is stably active in some queries and stably inactive in others. We observed no such cases in our experiments.}
Such instability variance poses a fundamental challenge to creating a compact and effective set of templates: a template that abstracts over a large number of unstable neurons is likely too imprecise to enable verification of the property of interest.  (Every unstable ReLU necessitates an over-approximation, introducing precision loss that compounds layer-by-layer, often rendering the final bounds too loose to be conclusive.)  And, since efficient template computation and matching is required for speedups, this issue cannot be sidestepped by larger template sets or more complex templates.

\subsection{Neuron Instability in Practice}\label{sec:limit-study}

Here, we describe a study that shows the limits of proof sharing via templates in practice and the relationship of neuron stability patterns to proof sharing potential.

\tightpara{Methodology} We used the MNIST and CIFAR-10 datasets in our study.  We considered six neural networks for each dataset, varying in training methods.  Alongside standard training, we tested the training methods
PGD~\cite{madry:17}, SABR~\cite{sabr}, TAPS~\cite{taps}, CURE~\cite{jiang2025universalcertifiedrobustnessmultinorm},
and CertifiedPatchDefense (CPD)~\cite{ChiangNAZSG20}. All the networks used in the limit study consist of 7 fully connected layers, with 200 neurons per layer.

We considered three perturbation types to generate verification queries from a common input image
that yield large number of queries and are amenable to be analyzed via abstract interpretation based verifiers: 
(1) random $L_0$ perturbations of 1 to 30 pixels generating 1000 queries, (2) localized $2 \times 2$ patch perturbations generating 729 queries for MNIST images and 1024 queries for CIFAR-10 images,
and (3) composite geometric transformations used in prior work on MNIST~\cite{balunovic2019certifying,10.1007/978-3-031-13185-1_7} including $\pm 2^\circ$ rotation, $\pm 10\%$ contrast and $\pm 1\%$ brightness changes split into $r$ perturbations.
We present $L_0$ perturbations as a representative case; results for patch and geometric perturbations, which exhibit similar trends, are deferred to \ref{sec:limit-study-appendix}.

For each generated query, we propagate the input region with DeepZ and record the abstract state $Q^i_k$ at each ReLU layer $k$ with $n_k$ neurons.
For a query set $\{Q^i_k\}_{i=1}^m$, \emph{Jointly Stable Neurons} at percentile $p \in [0,1]$ is the fraction of neurons stable in at least a $p$-fraction of the queries:
\begin{equation}
\mathrm{Jointly Stable Neurons}_p(\{Q^i_k\}_{i=1}^m) = \frac{\left|\left\{j : \dfrac{|\{i : j \notin \Unstable(Q^i_k)\}|}{m} \ge p\right\}\right|}{n_k}
\label{eq:jointly-stable-neurons}
\end{equation}
At $p=1$, this is the fraction of neurons outside the union of unstable neurons across all queries at that layer, i.e.,
$1 - |\bigcup_i \Unstable(Q^i_k)|/n_k$.   By \cref{eq:unstable-union-inclusion}, a smaller value means any template subsuming all queries must contain many unstable neurons.
A steep drop as $p$ increases indicates that instability is dispersed across queries, so templates covering an increasing fraction of queries must include more unstable neurons, limiting their effectiveness for proof sharing.

\begin{figure}[t!]
  \centering
  \begin{subfigure}[b]{\textwidth}
    \centering
    \includegraphics[width=\textwidth]{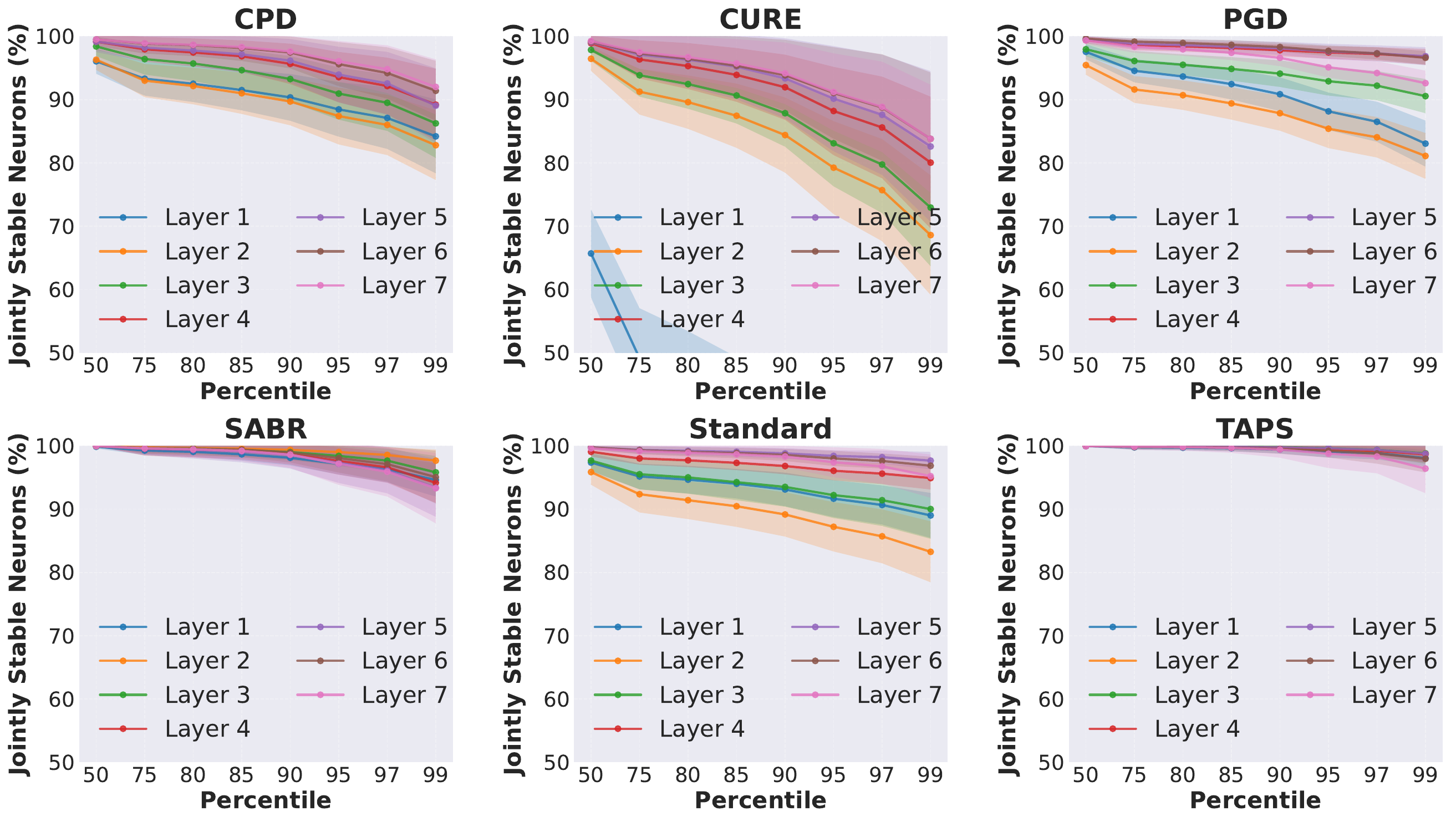}
    \caption{Jointly stable neurons}
  \end{subfigure}

  \vspace{0.5em}

  \begin{subfigure}[b]{\textwidth}
    \centering
    \includegraphics[width=\textwidth]{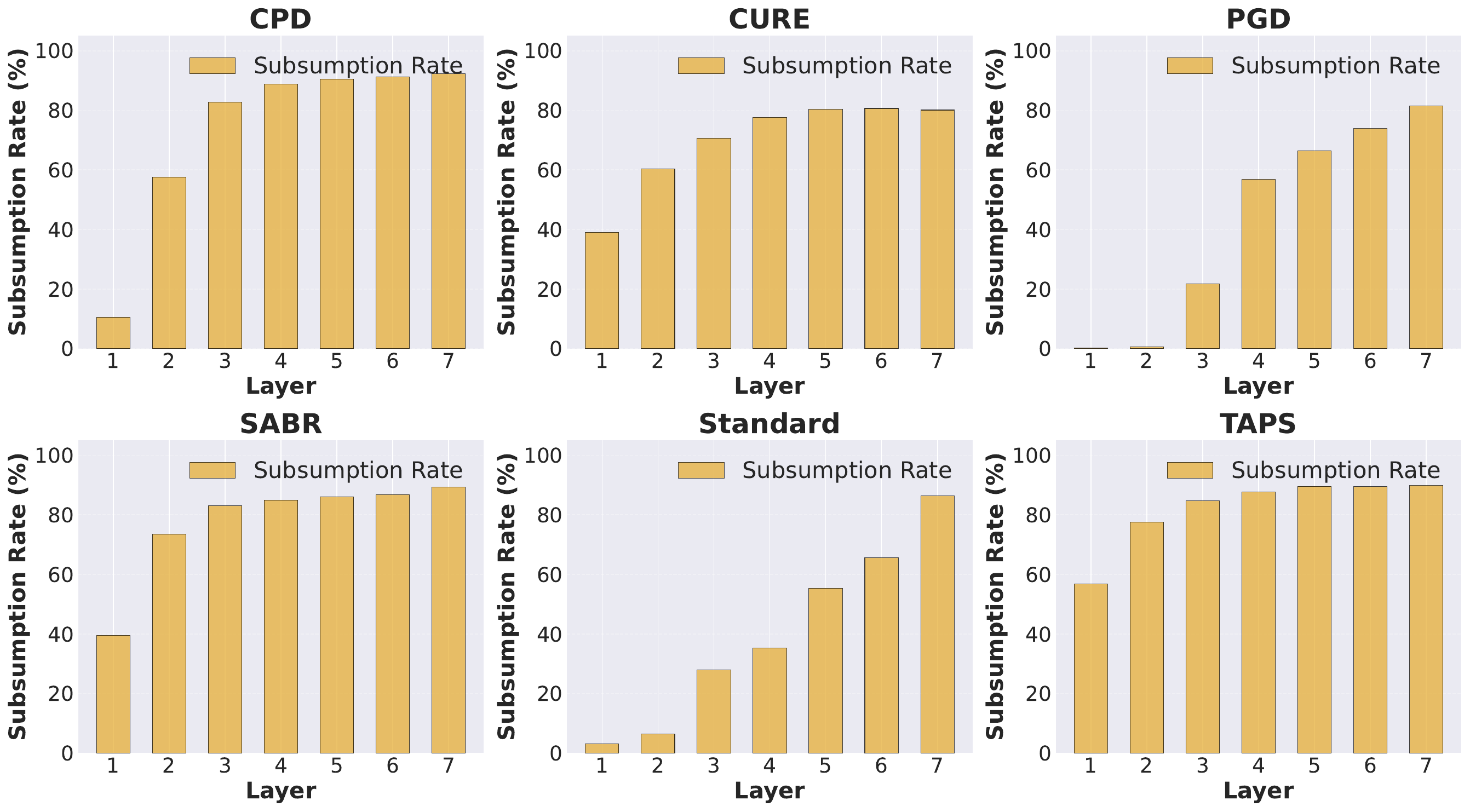}
    \caption{Template subsumption rates}
  \end{subfigure}

  \caption{Jointly stable neurons and template subsumption rates under random-$L_0$ perturbations in MNIST networks.}
  \label{fig:limit-study-mnist}
\end{figure}

\begin{figure}[t!]
  \centering
  \begin{subfigure}[b]{\textwidth}
    \centering
    \includegraphics[width=\textwidth]{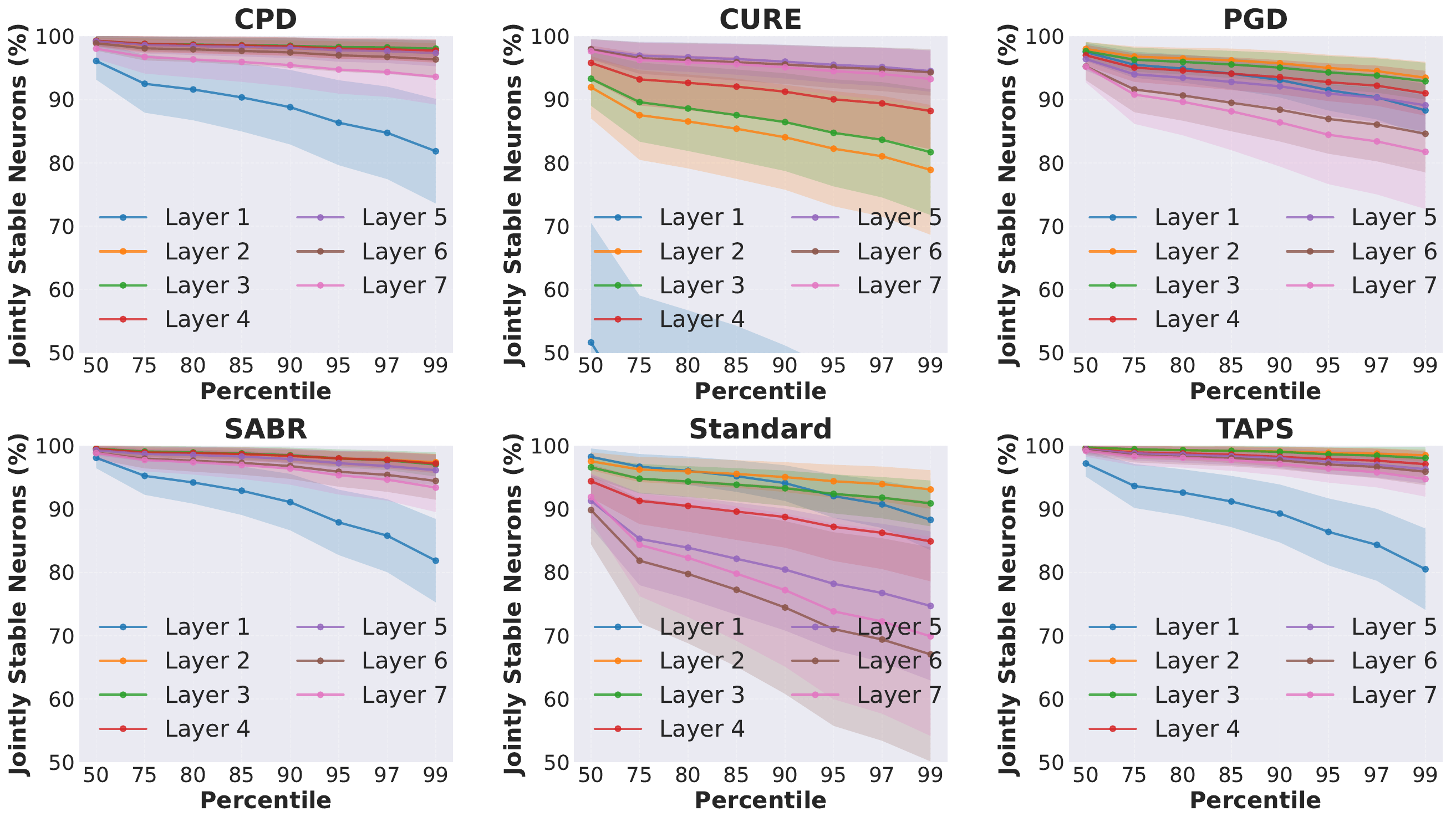}
    \caption{Jointly stable neurons}
  \end{subfigure}

  \vspace{0.5em}

  \begin{subfigure}[b]{\textwidth}
    \centering
    \includegraphics[width=\textwidth]{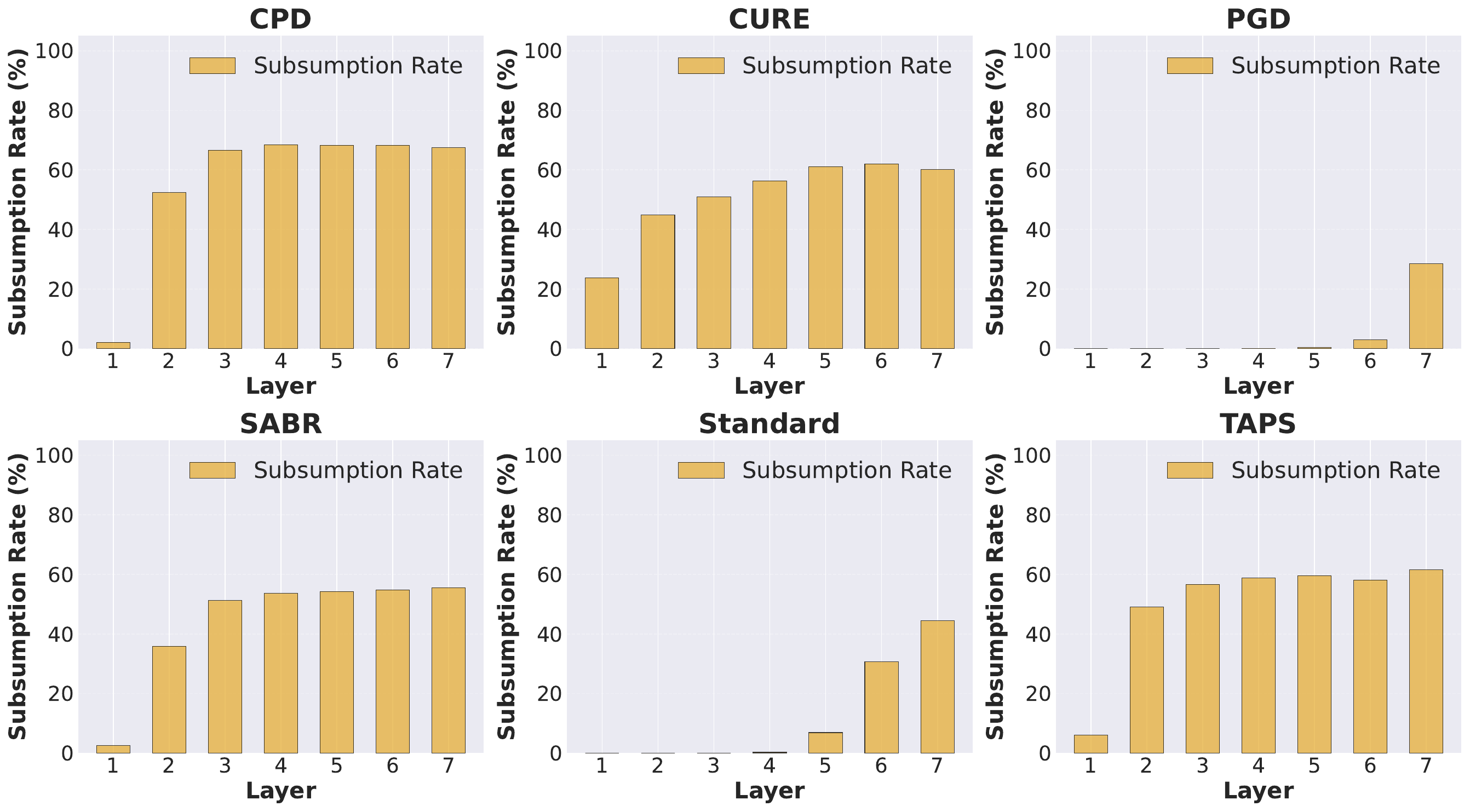}
    \caption{Template subsumption rates}
  \end{subfigure}

  \caption{Jointly stable neurons and template subsumption rates under random-$L_0$ perturbations in CIFAR-10 networks.}
  \label{fig:limit-study-cifar}
\end{figure}

In \cref{fig:limit-study-mnist,fig:limit-study-cifar}, we plot \textit{Jointly Stable Neurons} and template subsumption for all the ReLU layers of networks trained on MNIST and CIFAR-10, respectively,
using 100 input images per network; shaded bands indicate variability across images. The x-axis gives the percentile threshold $p$ expressed as a percentage, and the y-axis gives $\mathrm{Jointly Stable Neurons}_p(\{Q^i_k\}_{i=1}^m)$ as a percentage of neurons at that layer. 
We use one template generated under $L_\infty$ perturbation for subsumption checking; following the template generation technique described in \cref{template-generation}.

\tightpara{Subsumption Rates} First, focusing on the template subsumption rates in \cref{fig:limit-study-mnist,fig:limit-study-cifar}, we see a significant variance across datasets, layers, and training methods.  Consistent with prior observations~\cite{10.1145/3527319}, later layers consistently have higher subsumption rates, as by that point the networks tend to distill the semantic features of the input, enabling similar queries to yield similar abstract states.

Training method has a substantial impact on potential subsumption rates.
Networks trained without specific certifiable guarantees, the Standard and PGD trained models, exhibit generally lower subsumption rates: for MNIST this is most pronounced in the earlier layers, while for CIFAR-10 it persists across all layers. 
In contrast, networks trained with certified objectives (CURE, CPD, SABR, TAPS) are designed to satisfy worst-case guarantees against perturbations,
and thus tend to emphasize features that remain informative despite variations in the perturbations.  We see these behaviors reflected in generally higher subsumption rates for these training methods across both datasets.

However, even for certified-trained networks, the subsumption rates vary significantly across layers, and across datasets.  E.g., the subsumption rate at layer 2 on MNIST for TAPS roughly 80\% whereas for CPD its less than 60\%. This unpredictability led us to develop the profiling technique for automating template configuration described in \cref{sec:diagnostic}.

\tightpara{Joint Stability}
\Cref{fig:limit-study-mnist,fig:limit-study-cifar} show that in nearly all cases, greater joint stability across abstract states correlates with higher template subsumption, as expected.
For the majority of the cases, we observe that jointly stable neurons drop quickly as $p$ increases at the first layer for all networks.
Hence, if a single template were to subsume a majority of these abstract states, it must have a high number of unstable neurons, and hence would be unlikely to enable verification of the desired property.  The very low subsumption rates at the first layer confirm this predicted ineffectiveness.  Conversely, at later layers, the drop in jointly stable neurons is significantly less, indicating that the template will not have to accommodate high instability. And consequently, we observe higher subsumption at these layers.

We do observe some interesting exceptions to these trends.
On CIFAR-10 Standard and PGD in~\cref{fig:limit-study-cifar}, earlier layers exhibit high joint stability across percentile of queries, but subsumption remains low.
These cases indicate that while joint stability captures a key prerequisite for reuse, it does not guarantee that the resulting per-neuron bounds align tightly
enough across queries to be subsumed by a few templates. Even with similar joint stability, early-layer abstract states may be insufficiently distilled (or too high-variance in interval geometry).
Dataset complexity likely plays a role as well.  As MNIST is far less feature-rich than CIFAR-10, homogenic semantic structure may emerge earlier in the network, allowing early-layer templates to be useful.
For example, consider the TAPS and SABR-trained MNIST networks in~\cref{fig:limit-study-mnist}; unlike their CIFAR-10 counterparts, they exhibit some subsumption even in the first layer, and joint stability is maintained.

CURE on MNIST in~\cref{fig:limit-study-mnist} illustrates a nuanced case. Although joint stability drops at stricter percentile thresholds, the drop is not severe at the coverage levels relevant to the observed subsumption rates. In particular, subsumption rates remain around $60-80\%$ starting from layer 2, and the joint stability curves indicate that a large fraction of neurons remain jointly stable for a comparable fraction of queries. 
Thus, the low stability at the highest percentiles does not rule out reuse: it says that a single template is unlikely to cover nearly all queries without absorbing additional instability. Templates can still be effective when targeting a substantial, but not exhaustive, subset of the queries.

Overall, the results indicate that joint stability effectively explains subsumption rates for template-based proof sharing.%
\footnote{Similar trends hold for patch and geometric perturbations; results are deferred to~\ref{sec:limit-study-appendix}.}
The data show that the effectiveness of such proof sharing is highly configuration-dependent; a fixed policy for where and how to use templates will not generalize across these settings. In the next section, we describe a new adaptive strategy to automatically adapt to these varying conditions.

\section{Automatic Configuration of Templates}
\label{sec:diagnostic}

The practical efficacy of template-based proof sharing is critically dependent on a combination of factors: the training methodology,
the specific property under verification, the cardinality of the template set used, and the choice of layer for reuse.
An uninformed use of templates could actually degrade overall verification performance, a phenomenon we have observed in practice.

Prior work~\cite{10.1007/978-3-031-13185-1_7,10.1145/3527319} largely sidestepped this parameterization challenge, relying on manual,
empirical measurements to select a fixed set of $L_\infty$-masked templates, and up to two promising layers for template reuse.
While this methodology has demonstrated viability in constrained settings, manual tuning will not scale to a wide variety of real-world scenarios; the experiments of \cref{sec:limit-study} alone (including the results in \ref{sec:limit-study-appendix}) consider hundreds of combinations of training method, layer choice, and perturbation type.

Further, there have been significant advances in $L_0$ verification via techniques like Calzone~\cite{Shapira23} and CoverD~\cite{shapira2024boostingfewpixelrobustnessverification},
making $L_0$ verification increasingly practical but posing new challenges for proof sharing.
These methods have expanded verification settings from fixed-size $t$-pixel perturbations to arbitrary $k$-pixel perturbations where $k > t$.
Calzone~\cite{Shapira23} exploits the observation that robustness for any $k>t$ perturbed pixels implies robustness for every $t$ subset. It predicts a large $k$ via dynamic programming and sampling,
then uses covering designs to generate $k$-sized queries, submitted to a verifier, refining to smaller sets as needed. CoverD~\cite{shapira2024boostingfewpixelrobustnessverification} advances this covering-based approach by proposing covering verification designs,
an algorithm that selects between candidate coverings without constructing them but rather predicting their block-size distributions from closed-form mean/variance, then constructing the chosen covering on-the-fly.
The volume and structural complexity of queries generated by such methods can vary dramatically. Prior work~\cite{10.1145/3527319} demonstrates template reusability for random 3-pixel perturbations, but covering-design-based approaches involve verifying perturbations that may alter 100 or more pixels,
making them a challenging testbed for template reuse.

To improve the applicability of template reuse, we introduce a lightweight, \emph{a priori} profiling technique.
This technique provides a principled method to (i) identify and prune verification scenarios where using templates is futile,
and (ii) automatically select a set of advantageous layers and curate a compact set of templates to maximize anticipated performance gain.

\begin{algorithm}[t!]
\caption{\toolname{} Layer and Templates Selection}
\label{alg:diagnostic-refined}
\begin{algorithmic}[1]
\State \textbf{Input:} Network $N$, image $x$, candidate layers $\mathcal{L}_{\mathrm{cand}}$, sampling size $P$, candidate template counts $\mathcal{M}$.
\State \textbf{Output:} Set of pairs $(k, \mathcal{T}_k)$ of selected layers and their template sets

\vspace{0.25em}
\State $\mathsf{viable} \gets []$
\For{each $k\in \mathcal{L}_{\mathrm{cand}}$}
  \State $(m^\star,\rho_k^\star,\mathcal{T}_k^\star)\gets \text{None}$
  \For{each $m\in \mathcal{M}$}
    \State \emph{// Build $m$ templates at layer $k$ from image $x$}
    \State $\mathcal{T}_k^{(m)}\gets\emptyset$
    \State Generate $m$ number of $L_\infty$ $masks$
    \For{$i=1$ \textbf{to} $m$}
      \State Compute template $T_i$ at layer $k$ using ${mask}_i$; add $T_i$ to $\mathcal{T}_k^{(m)}$
    \EndFor
    \vspace{0.25em}
    \State \emph{// Estimate subsumption on the set of $P$  $L_0$ queries on image $x$}
    \State $c\_{{\rm hit}}\gets 0$
    \For{$j=1$ \textbf{to} $P$}
      \State Sample a $L_0$ query on $x$; compute $Q_j$ at layer $k$
        \For{$i=1$ \textbf{to} $m$}
            \If{$Q_j \sqsubseteq T_i$}
              \State $c\_{{\rm hit}}\gets c\_{{\rm hit}}+1$ \textbf{break}
            \EndIf
        \EndFor
    \EndFor
    \State $\rho_k(m)\gets c\_{{\rm hit}}/P$

    \State \emph{// Cost-model viability test}
    \If{$\rho_k(m) \ge \rho_k^{\min}(k,m,r)$}
      \State \emph{// Keep the best viable $(m,\rho)$ for this layer determined by subsumption}
      \State $(m^\star,\rho_k^\star,\mathcal{T}_k^\star)\gets (m,\rho_k(m),\mathcal{T}_k^{(m)})$
    \EndIf
  \EndFor
  \If{$(m^\star,\rho_k^\star,\mathcal{T}_k^\star)\neq\text{None}$}
    \State Append $(k,\rho_k^\star,m^\star,\mathcal{T}_k^\star)$ to $\mathsf{viable}$
  \EndIf
\EndFor
\State \textbf{return} $\{(k,\mathcal{T}_k^\star) : (k,\cdot,\cdot,\mathcal{T}_k^\star,\cdot)\in \mathsf{viable}\}$
\end{algorithmic}
\end{algorithm}

\subsection{Layer Selection and Template Set Refinement}

In practice, not all layers are equally suitable for template reuse.  Early layers may exhibit low subsumption rates (see \cref{sec:limit-study}),
while very late layers offer limited potential savings due to the small residual depth. And, the number of templates must be balanced against lookup costs, as larger template sets increase both generation and matching overheads.
Identification of a set of layers and template sets that can exhibit the highest potential for subsumption is governed by the performance model from \cref{sec:template-model}. 
We leverage this model in \cref{alg:diagnostic-refined} to guide the selection of layers and templates for \toolname{}. The algorithm takes as input a network,
an input image, a set of candidate reuse layers $\mathcal{L}_{\mathrm{cand}}$, sampling size $P$, and a set of candidate template counts $\mathcal{M}$. 
It evaluates each layer-template configuration $(k,m)$ to assess its viability for template reuse, and outputs the set of viable $(k,m)$ configurations.

At each layer, to achieve a speedup \(S>1\), the subsumption fraction \(\rho_k\) must exceed a minimum threshold \(\rho_k^{\min}\) that depends on the amortized overheads and the residual network depth,
\(L-k\). This threshold is derived from \cref{eq:speedup-constlayer}:
\[
\rho_k^{\min}
\;=\;
\Bigl(\lambda\,\tfrac{m}{r} + \tfrac{\eta\,m}{L\,\nu}\Bigr)\,\frac{L}{\,L-k\,}
\qquad (k<L).
\]

Layers with subsumption fractions below a minimum threshold $\rho_k^{\min}$ will not yield sufficient savings to
justify the overhead of matching. In \cref{alg:diagnostic-refined}, \toolname{} evaluates each candidate layer $k \in \mathcal{L}_{\mathrm{cand}}$ and each candidate template count $m \in \mathcal{M}$, 
using an estimation of the subsumption fraction $\rho_k(m)$ by sampling a set of diverse $L_0$ queries and computing the fraction subsumed by a set of $m$ templates. 
A layer-template configuration $(k, m)$ is viable if it meets the threshold $\rho_k(m) \ge \rho_k^{\min}(k,m,r)$.

\[
\mathcal{L}_{\mathrm{viable}}
=
\{k \in \mathcal{L}_{\mathrm{cand}} : \rho_k \ge \rho_k^{\min}\},
\]

If no layer-template configuration meets the viability threshold, the algorithm returns an empty set and verification proceeds without templates.
This design ensures that template-based verification is applied only when empirical evidence suggests a net performance gain, 
gracefully degrading to standard verification when template reuse is unprofitable.

\section{Methodology}
\label{label:methodology}

In this section, we describe the methodology used to experimentally validate the effectiveness of \toolname{}.

\subsection{Configuration}
We instantiate \toolname{} with verification queries generated by Calzone~\cite{Shapira23} and CoverD~\cite{shapira2024boostingfewpixelrobustnessverification}, two state-of-the-art $L_0$ robustness verification frameworks.
Both frameworks generate verification queries using covering designs to efficiently explore the space of possible pixel perturbations.
While the original Calzone and CoverD frameworks fall back to complete solvers when search tree over pixel sets is exhausted and reaches leaf nodes, we exclusively use the incomplete verifier for all queries.
This isolates the performance impact of template reuse as an improvement on the core abstract interpretation backend.

\begin{wrapfigure}[17]{R}{0.48\textwidth}
  \centering
  \includegraphics[width=0.48\textwidth]{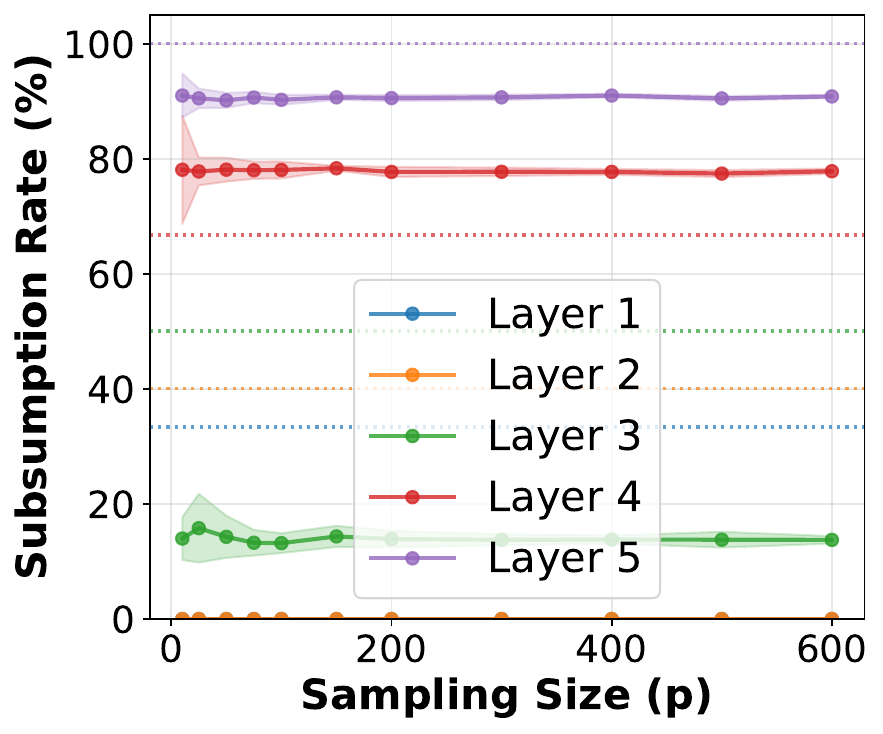}
  \caption{Sampling size vs. subsumption rate estimates for MNIST PGD with 4 templates.}
  \label{fig:pgd-mnist-p}
\end{wrapfigure}

To create a comprehensive yet tractable evaluation, we generate a total of 750,000 verification queries per network. This total was derived from 10 images per network. 
We instantiate the workload for $t$-pixel perturbation tasks with $t=3$ with Calzone and CoverD. They employ covering-design to generate query sets whose perturbed
pixel counts, $k > t$, can vary significantly, producing a heterogeneous set of verification problems. For each image, we cap the workload at 75,000 queries. To ensure consistent query set across different methods compared, we cache the sampling phase of Calzone/CoverD and 
reuse across all evaluations.

To estimate the subsumption rate $\rho_k$ at each candidate layer, we perform a brief profiling run on a small sample of $P=200$ queries per image.
Across networks and layers, subsumption estimates vary substantially for small samples (up to $\sim$150 queries),
but stabilize around $\sim$200 queries. Importantly, the threshold-crossing decision is not sensitive to sampling variability beyond this point (e.g., \cref{fig:pgd-mnist-p}).
We give more data justifying our choice of $P$ in \mbox{\ref{sec:implementation-appendix}}.

To balance template diversity against lookup costs, we consider a set of candidate template set sizes of, $\mathcal{M} = \{1, 2, 4\}$. $\mathcal{L}_{\mathrm{cand}}$ includes the layers 
where the minimum subsumption fraction $\rho_k^{\min}$ is less than or equal to 1 for at least one $m \in \mathcal{M}$, ensuring that only layers with potential for speedup are considered.

We evaluate \toolname{} on fully-connected and convolutional networks trained on MNIST and CIFAR-10 (summarized in \cref{tab:networks}). The suite includes standard,
robust, and certified-trained models of varied sizes used in prior work~\cite{Shapira23,shapira2024boostingfewpixelrobustnessverification,10.1007/978-3-031-13185-1_7,10.1145/3527319}. As we observed that SABR and TAPS trained networks exhibited substantially lower verifiability under Calzone/CoverD-generated $L_0$ perturbation queries with our incomplete verifier domains,
we exclude them from the final evaluation set.

We base our evaluation using \textsc{DeepZ}~\cite{singh2018fast,10.1007/978-3-031-13185-1_7} and {Box}~\cite{mirman2018differentiable} abstract domains, and compare \toolname{} against two baseline techniques for both the domains.
The first one is a standard verification approach that does not attempt template-based proof sharing.
The other one uses a static, hand-picked reuse strategy described in prior work. It always enables template reuse with one template ($m=1$) at fixed, pre-selected layers (layers 2 and 3~\cite{10.1007/978-3-031-13185-1_7,10.1145/3527319}). 
This allows us to measure the benefit of \toolname's layer and template selection strategy over the state-of-the-art policy.

\begin{table}[H]
\vspace{-.15in}
\caption{Networks used in the \toolname{} evaluation.}
\label{tab:networks}
\centering
\begin{tabular}{llcc}
\toprule
Dataset & Model & Architecture & \#Neurons \\
\midrule
\multirow{6}{*}{MNIST, CIFAR-10} & CPDMed & 7$\times$200 linear layers & 1400 \\
& CPDBig & 9$\times$500 linear layers & 4500 \\
& CURE & 9$\times$500 linear layers & 4500 \\
& PGD & 6$\times$500 linear layers & 3000 \\
& Standard & 6$\times$500 linear layers & 3000 \\
& CPDConv & 4 Conv layers, 4 linear layers & 8448 \\
\bottomrule
\end{tabular}
\vspace{-.1in}
\end{table}

\subsection{Implementation and Environment}\label{sec:implementation-and-environment}

\toolname{} is implemented in Python and extends the \textsc{DeepZ}~\cite{singh2018fast,10.1007/978-3-031-13185-1_7}
verifier. We integrated the query generation algorithms from Calzone~\cite{Shapira23} and CoverD~\cite{shapira2024boostingfewpixelrobustnessverification} to
create our verification tasks. All experiments were conducted on a Google Cloud Platform
instance equipped with a 12-core Intel Cascade Lake CPU, 85GB of RAM, and an NVIDIA A100 GPU with 40GB of memory.

\section{Experimental Evaluation}
\label{sec:evaluation}

We conducted an empirical evaluation to validate the efficacy of \toolname. The evaluation showed that the naive reuse of existing policy for template-based proof sharing can be detrimental to performance,
and that our automated, profile-guided approach can both mitigate this pitfall and maximize gains. Our investigation was structured around the following research questions:

\begin{itemize}
\item  RQ1: What is the impact of the previous fixed-layer template reuse strategy on end-to-end verification time for $L_0$ verification queries?
\item  RQ2: In cases where the previous strategy reduces verification performance, can \toolname's ability to detect when template reuse is futile mitigate the performance reduction?  What is the overhead of the profiling process in this scenario?
\item  RQ3: In cases where the extant reuse strategy improves verification performance, what further performance gains are obtained from \toolname's automated selection of layers and template count?
\item  RQ4: Does \toolname generalize across the abstract domains used by incomplete verifiers, in settings where template reuse is applicable?
\end{itemize}
We answer RQ1 through RQ4 respectively in \cref{sec:eval-extant-reuse-policy,sec:eval-futility,sec:eval-favorable,sec:eval-generality}.  Finally, we discuss limitations in \cref{sec:limitations}.

\begin{table*}[t]
\centering
\caption{Calzone results (750K queries, 10 images, t=3) with DeepZ and Box domains. V=Verified(\%), T=Baseline (No Template) Time in seconds, Sub=Subsumption
rate (\%), Spd=Speedup, E=Extant Template Reuse Strategy, F=\toolname.}
\label{tab:calzone-combined-cols}

\vspace{1mm}
(a) MNIST
\label{tab:calzone-combined-cols-mnist}

\resizebox{\textwidth}{!}{%
\normalsize
\setlength{\tabcolsep}{3pt}
\begin{tabular}{@{}l|cccccc|cccccc@{}}
\toprule
& \multicolumn{6}{c|}{DeepZ} & \multicolumn{6}{c}{Box} \\
\cmidrule(lr){2-7} \cmidrule(lr){8-13}
Net & V & T & \multicolumn{2}{c}{Sub} & \multicolumn{2}{c|}{Spd} & V & T & \multicolumn{2}{c}{Sub} & \multicolumn{2}{c}{Spd} \\
& (\%) & (s) & E & F & E & F & (\%) & (s) & E & F & E & F \\
\midrule
CPDMed  & 92.4 & 5293 & 57.8 & \textbf{85.2} & 1.28 & \textbf{1.41} & 94.6 & 4310 & 28.7 & \textbf{92.0} & 1.02 & \textbf{1.40} \\
CPDBig  & 88.4 & 6309 & 87.0 & \textbf{98.6} & 1.91 & \textbf{2.10} & 92.8 & 4966 & 62.9 & \textbf{98.7} & 1.39 & \textbf{2.00} \\
CPDConv & 84.5 & 4781 & 68.4 & \textbf{72.0} & \textbf{1.08} & 1.06 & 84.5 & 8752 & 52.6 & \textbf{68.4} & 0.99 & \textbf{1.00} \\
CURE    & 44.5 & 4704 & 43.7 & \textbf{85.3} & 1.12 & \textbf{1.24} & 31.5 & 2890 & 33.0 & \textbf{82.9} & 1.03 & \textbf{1.10} \\
PGD     & 92.6 & 5166 & 0.5 & \textbf{88.7} & 0.91 & \textbf{1.17} & 86.7 & 4247 & 0.0 & \textbf{40.5} & 0.87 & \textbf{1.03} \\
Std     & 96.5 & 4925 & 56.0 & \textbf{90.2} & 1.21 & \textbf{1.34} & 89.7 & 4265 & \textbf{0.0} & \textbf{0.0} & 0.87 & \textbf{0.98} \\
\bottomrule
\end{tabular}%
}

\vspace{3mm}
(b) CIFAR-10
\label{tab:calzone-combined-cols-cifar}

\resizebox{\textwidth}{!}{%
\normalsize
\setlength{\tabcolsep}{3pt}
\begin{tabular}{@{}l|cccccc|cccccc@{}}
\toprule
& \multicolumn{6}{c|}{DeepZ} & \multicolumn{6}{c}{Box} \\
\cmidrule(lr){2-7} \cmidrule(lr){8-13}
Net & V & T & \multicolumn{2}{c}{Sub} & \multicolumn{2}{c|}{Spd} & V & T & \multicolumn{2}{c}{Sub} & \multicolumn{2}{c}{Spd} \\
& (\%) & (s) & E & F & E & F & (\%) & (s) & E & F & E & F \\
\midrule
CPDMed  & 97.6 & 5997 & 43.1 & \textbf{60.9} & 1.20 & \textbf{1.25} & 98.9 & 4935 & 33.6 & \textbf{55.1} & 1.13 & \textbf{1.34} \\
CPDBig  & 95.2 & 7406 & 46.3 & \textbf{68.4} & 1.31 & \textbf{1.45} & 97.7 & 6032 & 62.8 & \textbf{79.4} & 1.50 & \textbf{1.80} \\
CPDConv & 85.3 & 3451 & 58.5 & \textbf{84.5} & 1.01 & \textbf{1.08} & 96.8 & 6721 & 67.7 & \textbf{88.6} & 1.01 & \textbf{1.03} \\
CURE    & 94.8 & 7161 & 3.5 & \textbf{31.5} & 0.94 & \textbf{1.02} & 82.7 & 6031 & 0.5 & \textbf{9.0} & 0.90 & \textbf{1.01} \\
PGD     & 58.1 & 5227 & 0.0 & \textbf{10.2} & 0.92 & \textbf{1.05} & 14.4 & 4298 & \textbf{0.0} & \textbf{0.0} & 0.86 & \textbf{0.99} \\
Std     & 54.3 & 5071 & \textbf{0.0} & \textbf{0.0} & 0.90 & \textbf{0.99} & 13.0 & 4328 & \textbf{0.0} & \textbf{0.0} & 0.88 & \textbf{0.99} \\
\bottomrule
\end{tabular}%
}
\vspace{-2mm}
\end{table*}

\begin{table*}[t]
\centering
\caption{CoverD results (750K queries, 10 images, t=3) with DeepZ and Box domains. V=Verified(\%), T=Baseline (No Template) Time in seconds, Sub=Subsumption
rate (\%), Spd=Speedup, E=Extant Template Reuse Strategy, F=\toolname.}
\label{tab:coverd-combined-cols}

\vspace{1mm}
(a) MNIST
\label{tab:coverd-combined-cols-mnist}

\resizebox{\textwidth}{!}{%
\normalsize
\setlength{\tabcolsep}{3pt}
\begin{tabular}{@{}l|cccccc|cccccc@{}}
\toprule
& \multicolumn{6}{c|}{DeepZ} & \multicolumn{6}{c}{Box} \\
\cmidrule(lr){2-7} \cmidrule(lr){8-13}
Net & V & T & \multicolumn{2}{c}{Sub} & \multicolumn{2}{c|}{Spd} & V & T & \multicolumn{2}{c}{Sub} & \multicolumn{2}{c}{Spd} \\
& (\%) & (s) & E & F & E & F & (\%) & (s) & E & F & E & F \\
\midrule
CPDMed  & 89.9 & 4768 & 59.7 & \textbf{85.8} & 1.24 & \textbf{1.35} & 91.7 & 3850 & 32.2 & \textbf{91.2} & 1.07 & \textbf{1.45} \\
CPDBig  & 87.5 & 5842 & 89.2 & \textbf{98.7} & 1.88 & \textbf{2.04} & 90.8 & 4916 & 60.1 & \textbf{98.8} & 1.33 & \textbf{2.00} \\
CPDConv & 84.9 & 4530 & 69.3 & \textbf{72.3} & \textbf{1.02} & \textbf{1.02} & 84.3 & 8602 & 53.7 & \textbf{70.1} & 1.04 & \textbf{1.05} \\
CURE    & 43.3 & 4321 & 53.4 & \textbf{89.8} & 1.12 & \textbf{1.24} & 30.3 & 2925 & 41.8 & \textbf{78.8} & 1.08 & \textbf{1.14} \\
PGD     & 89.4 & 4896 & 0.4 & \textbf{86.5} & 0.89 & \textbf{1.09} & 79.5 & 4323 & 0.0 & \textbf{48.4} & 0.86 & \textbf{1.02} \\
Std     & 96.9 & 4716 & 62.8 & \textbf{93.5} & 1.25 & \textbf{1.34} & 89.4 & 4429 & \textbf{0.0} & \textbf{0.0} & 0.88 & \textbf{0.96} \\
\bottomrule
\end{tabular}%
}

\vspace{3mm}
(b) CIFAR-10
\label{tab:coverd-combined-cols-cifar}

\resizebox{\textwidth}{!}{%
\normalsize
\setlength{\tabcolsep}{3pt}
\begin{tabular}{@{}l|cccccc|cccccc@{}}
\toprule
& \multicolumn{6}{c|}{DeepZ} & \multicolumn{6}{c}{Box} \\
\cmidrule(lr){2-7} \cmidrule(lr){8-13}
Net & V & T & \multicolumn{2}{c}{Sub} & \multicolumn{2}{c|}{Spd} & V & T & \multicolumn{2}{c}{Sub} & \multicolumn{2}{c}{Spd} \\
& (\%) & (s) & E & F & E & F & (\%) & (s) & E & F & E & F \\
\midrule
CPDMed  & 96.5 & 6047 & 47.3 & \textbf{68.5} & 1.21 & \textbf{1.29} & 98.4 & 4858 & 37.6 & \textbf{58.1} & 1.12 & \textbf{1.36} \\
CPDBig  & 95.4 & 7495 & 54.3 & \textbf{75.4} & 1.37 & \textbf{1.53} & 97.2 & 6314 & 63.4 & \textbf{81.1} & 1.51 & \textbf{1.86} \\
CPDConv & 84.4 & 3376 & 60.8 & \textbf{84.1} & 1.02 & \textbf{1.04} & 94.6 & 5508 & 75.0 & \textbf{91.7} & 1.00 & \textbf{1.01} \\
CURE    & 95.9 & 7174 & 6.4 & \textbf{40.4} & 0.93 & \textbf{1.05} & 83.9 & 6203 & 0.1 & \textbf{8.3} & 0.91 & \textbf{1.02} \\
PGD     & 58.1 & 5444 & \textbf{0.0} & \textbf{0.0} & 0.92 & \textbf{0.99} & 16.2 & 4645 & \textbf{0.0} & \textbf{0.0} & 0.88 & \textbf{1.00} \\
Std     & 51.1 & 5362 & \textbf{0.0} & \textbf{0.0} & 0.91 & \textbf{1.00} & 13.1 & 4614 & \textbf{0.0} & \textbf{0.0} & 0.87 & \textbf{0.99} \\
\bottomrule
\end{tabular}%
}
\vspace{-2mm}
\end{table*}

\subsection{Performance of Extant Template Reuse Policy}\label{sec:eval-extant-reuse-policy}

Extant reuse yields substantial speedups on networks where subsumption is moderately high, but consistently slows verification when subsumption is very low.
From \cref{tab:calzone-combined-cols,tab:coverd-combined-cols,fig:combined-speedup-eval}, we observe
two distinct behaviors,
  \noindent \textit{When reuse is viable,} extant reuse achieves speedup ranging $1.01\times-1.91\times$ on Calzone,
and $1.02\times-1.88\times$ in CoverD.
  \noindent \textit{When reuse is futile,} extant reuse slows down verification by $1\%-14\%$ on Calzone and $7\%-14\%$ on CoverD.
This performance degradation arises because the overhead of template matching outweighs the negligible savings from subsumption. 
The performance degradation in these cases is predicted by the template reuse cost model in \cref{sec:template-model}, and by the measures of
joint stability from~\cref{eq:jointly-stable-neurons} presented in~\cref{sec:proof-sharing-potential}, particularly in~\cref{fig:limit-study-mnist,fig:limit-study-cifar}.

\subsection{Efficacy of Automated Futility Detection}\label{sec:eval-futility}
In scenarios where template reuse is detrimental, \toolname's automated futility detection effectively avoids slowdowns.
Automated futility detection in \toolname can identify these scenarios and disable reuse, preventing slowdowns. Our profiling step figures out quickly that no layers are viable for reuse, returning empty layer-template configuration set (\cref{alg:diagnostic-refined}).
This avoids the added overhead of template matching, resulting in verification times comparable to the baseline without reuse, as seen
in \cref{tab:calzone-combined-cols,tab:coverd-combined-cols}.
The remaining slowdown stems from template creation and matching against 200 $L_0$ queries per image during profiling,
but this cost is negligible compared to the 75,000 verification queries per image.
\toolname only incurs an average slowdown of $1.6\%$ (ranging over $1\% - 4\%$), a significant improvement over the $10.1\%$ (ranging over $1\% - 14\%$) average slowdown observed with extant template reuse. This 
is demonstrated across both Calzone and CoverD in \cref{fig:combined-speedup-eval}, where \toolname consistently outperforms extant reuse in futile scenarios.

\subsection{Advantage in Favorable Scenarios}\label{sec:eval-favorable}
\label{subsec:eval-favorable} 
In configurations where template reuse is effective, \toolname{}'s automated selection of reuse layers and
template counts improves over extant reuse, ultimately achieving an average of \speedup{} (ranging over $0.98\times - 1.5\times$) performance gain overall.

\begin{figure}[t]
  \centering
  \includegraphics[width=\textwidth]{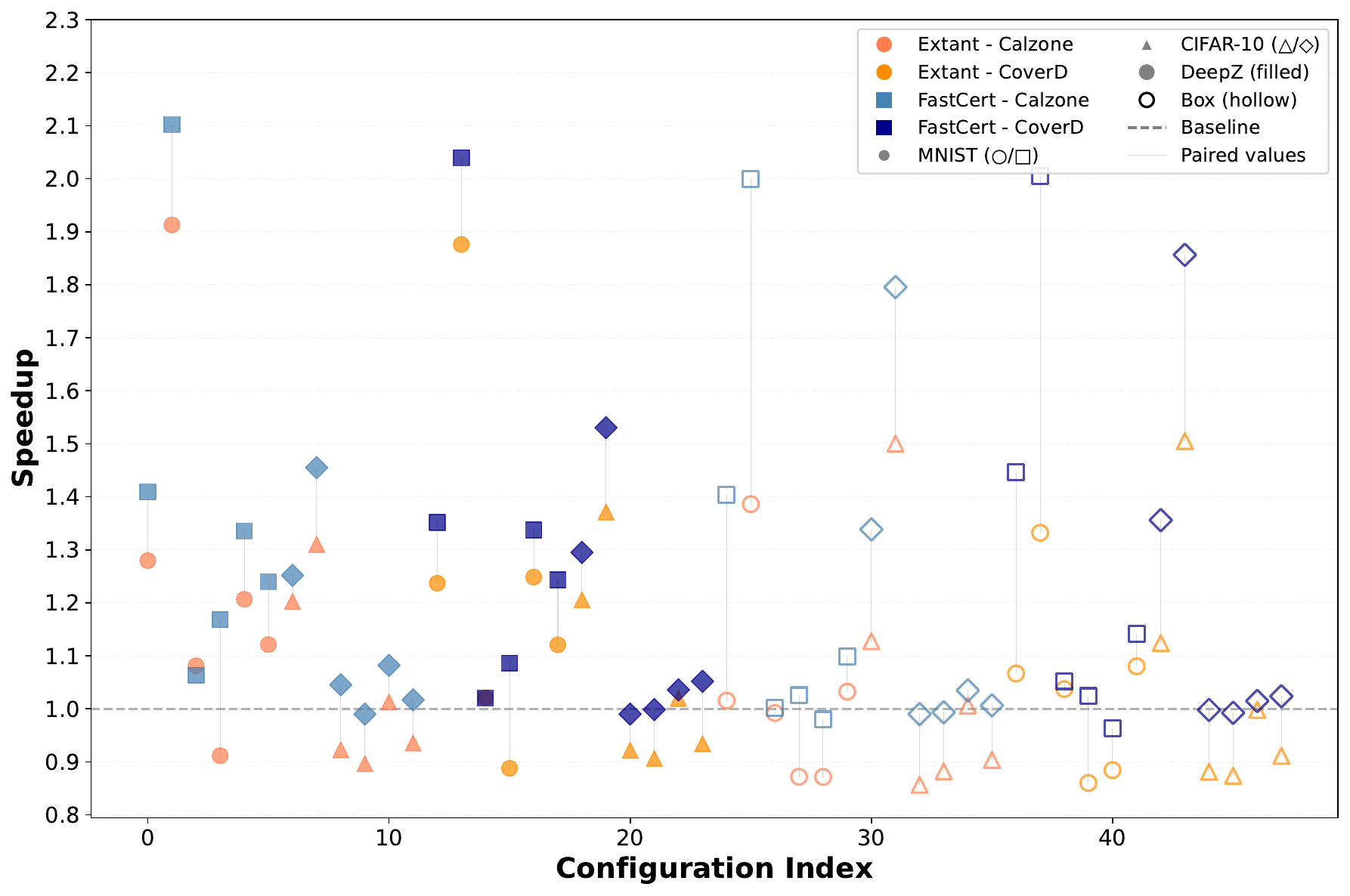}
  \caption{Combined speedup results across all 48 configurations: \toolname{} outperforms extant reuse in 46, ties in one, and is slightly slower in one; the overall improvement is statistically significant according to a paired two-sided Wilcoxon signed-rank test ($W=4$, $p=3.1\times10^{-9}$).}
  \label{fig:combined-speedup-eval}
\end{figure}

Overall, \toolname{} achieves speedups of up to $2.10\times$ on MNIST networks (e.g., \textsc{CPDBig}) and up to $1.86\times$ on CIFAR-10 networks 
(e.g., \textsc{CPDBig}) compared to no template reuse~(\cref{tab:calzone-combined-cols,tab:coverd-combined-cols,fig:combined-speedup-eval}). These peak improvements arise because \toolname{} (i) identifies layers with
higher template subsumption and (ii) chooses template counts that increase hit rate while controlling template-matching overhead.
For example, for a representative MNIST input on \textsc{CPDMed} for Calzone, \toolname{} applies two templates each at layers~2 and 3, and one template each at
layers~4 and~5. In contrast, extant reuse relies on a static, manually chosen configuration; \toolname{} adapts reuse layers and template counts to the subsumption behavior of each network and dataset.

In CPDConv the gains are minor since reuse occurs only at late linear layers followed by convolutional layers,
which constrains subsumption potential and leaves little room for improvement~\cite{10.1145/3527319}.

\subsection{Generality Across Abstract Domains}\label{sec:eval-generality}
We evaluate \toolname{} with both DeepZ and Box, two abstract domains that are compatible with template reuse as identified in prior work~\cite{10.1007/978-3-031-13185-1_7}.
We also experimented with another widely used abstract domain, Polyhedra (with DeepPoly~\cite{singh2019abstract}), but found them impractically slow at our evaluation scale: for example, on
\textsc{CPDMed} (MNIST) under CoVerD with 750k queries and no template reuse, DeepPoly is $8.08\times$ slower than DeepZ and $9.54\times$ slower than Box.
Accordingly, we focus our evaluation on DeepZ and Box.

From \cref{fig:combined-speedup-eval}, we observe that \toolname achieves almost similar speedup trends in both domains, with slightly better performance in Box. Specifically, \toolname achieves 
$0.98\times-1.29\times$ speedup over extant reuse in DeepZ with an average of $1.10\times$, and $1.01\times-1.5\times$ speedup over extant reuse in Box domain with an average of $1.17\times$ across the tools and datasets.
This shows that \toolname generalizes well across both the abstract domains.

\subsection{Limitations}\label{sec:limitations}
We study template reuse for robustness verification in the incomplete-verifier setting, and instantiate \toolname{}
with the incomplete verification backends of Calzone and CoVerD. While these tools may ultimately invoke a complete procedure
when the abstract domain cannot discharge a query at the leaf levels, \toolname{} operates only on the incomplete backend and is
agnostic to any subsequent complete verification step.
Our work is limited to verification queries generated from a single input image (local robustness).
Extending template reuse across multiple inputs; for example, to support global robustness properties or batched/multi-input settings, 
is another promising direction. The performance model used in this work employs a constant per layer cost to derive viability threshold.
Verification cost can depend on the layer type, size and on the abstract domain; although our end-to-end measurements capture this variation,
extending \toolname{} with a layer-aware cost model may further improve its configuration decisions.

\section{Related Work}
\label{sec:related-work}

\textbf{DNN Verification.}
Existing DNN verifiers can be broadly classified as complete or incomplete~\cite{gehr2018ai2,singh2018fast,singh2019abstract,singh2019beyond,zhang2018crown,Lirpa:20,yang2023provable,DBLP:conf/nips/SalmanY0HZ19,Shapira23,shapira2024boostingfewpixelrobustnessverification}.
Incomplete methods trade some precision for scalability, enabling the verification of larger and more complex networks that complete methods struggle to handle.
Most of the prior works on DNN verification is focused on $L_\infty$ verification~\cite{gehr2018ai2,singh2018fast,singh2019abstract}. 
Unlike $L_\infty$ robustness, $L_0$ robustness induces a discrete, combinatorial, non-convex perturbation space, making scalable certification substantially more challenging.
Calzone~\cite{Shapira23} and CoverD~\cite{shapira2024boostingfewpixelrobustnessverification} address this challenge with covering designs, reducing $t$-pixel robustness to large collections of variable-size $k$-pixel queries with $k>t$; in practice, this can produce tens to hundreds of thousands of related queries per image.
More recent work~\cite{shapira2026tightrobustnesscertificationconvex} advances this line by characterizing the convex hull of an $L_0$ ball as the intersection of its bounding box and a scaled $L_1$-like polytope.
Prior template-reuse work demonstrates benefits on much smaller fixed-size $L_0$ workloads, but none study whether template reuse remains effective in covering-design setup.

\noindent\textbf{Incremental Verification.}
Incremental verification has improved the scalability of traditional program verification to an industrial scale \cite{10.1145/2465449.2465456,Lachnech-etal:TACAS01,DBLP:conf/lics/OHearn18,DBLP:conf/pldi/0002CS21}. 
Incremental program analysis tasks reuse partial results \cite{5306334}, constraints \cite{10.1145/2393596.2393665}, and precision information \cite{10.1145/2491411.2491429} from previous runs for faster analysis of individual commits.  
Program changes typically affect a small, localized part of the code, whereas DNN updates modify weights across many layers without altering control flow.
This key difference makes incremental DNN verification a distinct challenge in comparison to program analysis. 

Similarly, many recent works~\cite{10.1145/3527319,10.1007/978-3-031-13185-1_7,10.1145/3591299,ugare2024incremental,10.1145/3720417} have successfully used incremental verification for improving the efficiency of DNN verification.
These works include incremental complete verification~\cite{10.1145/3591299,10.1145/3720417}, incremental probabilistic verification~\cite{ugare2024incremental} and incremental incomplete verification~\cite{10.1145/3527319,10.1007/978-3-031-13185-1_7}.
Our work extends this line by systematically analyzing when and how verification effort can be reused effectively, providing both theoretical and automated support for template reuse.

Wei et al.~\cite{DBLP:journals/corr/abs-2106-12732} considers incremental incomplete verification of relatively small DNNs with last-layer perturbation.
Fischer et al.~\cite{10.1007/978-3-031-13185-1_7} introduced proof transfer to reuse verification effort across different input specifications of a single network, achieving up to 2.9x reduction in verification cost but with limited applicability to convolutional layers.
Complementarily, Ugare et al.~\cite{10.1145/3527319} extended this idea to transfer proofs across approximate networks by introducing a template transformation technique that enables sound and efficient verification for networks with convolutional architectures.
Our work builds on these ideas by conducting a limit study of template effectiveness, identifying conditions where template reuse is counterproductive, and proposing an automated system that decides when, where, and how to apply template reuse efficiently.

\section{Conclusions}\label{sec:conclusions}

We have presented a study of the limits of template-based proof sharing for neural network verification.  We showed that certain combinations of networks, training methods, and datasets are fundamentally unsuited to template-based speedups, due to excessive variance in jointly stable neurons.  Then, we gave a new technique for automatically determining how best to apply templates to a verification task, including automatic detection of when templates should not be used.  We implemented our technique in a tool \toolname, and an experimental evaluation showed that it improves over a state-of-the-art template technique, providing greater speedup in cases where templates work well and significantly reducing slowdown in cases where they are unsuitable.

\subsubsection*{Artifact Availability.}
The artifact accompanying this paper, including the \toolname{} source code,
networks, and experimental scripts, is available on Zenodo
(DOI: \href{https://doi.org/10.5281/zenodo.21314767}{10.5281/zenodo.21314767}).

\subsubsection*{Acknowledgments.}

This research was supported in part by the National Science Foundation under grants CCF-2238079, CCF-2313028, CCF-2223825, CCF-2223826, a gift from Oracle Labs, and a Google Research Award. The views expressed herein are those of the authors and do not necessarily reflect those of our funders.

\bibliographystyle{splncs04}
\bibliography{refs}

@article{10.1145/3527319,
author = {Ugare, Shubham and Singh, Gagandeep and Misailovic, Sasa},
title = {Proof transfer for fast certification of multiple approximate neural networks},
year = {2022},
issue_date = {April 2022},
publisher = {Association for Computing Machinery},
address = {New York, NY, USA},
volume = {6},
number = {OOPSLA1},
url = {https://doi.org/10.1145/3527319},
doi = {10.1145/3527319},
journal = {Proc. ACM Program. Lang.},
month = apr,
articleno = {75},
numpages = {29}
}

@article{DBLP:journals/corr/abs-2106-12732,
  author    = {Tianhao Wei and
               Changliu Liu},
  title     = {Online Verification of Deep Neural Networks under Domain or Weight
               Shift},
  journal   = {CoRR},
  volume    = {abs/2106.12732},
  year      = {2021},
  _url       = {https://arxiv.org/abs/2106.12732},
  eprinttype = {arXiv},
  eprint    = {2106.12732},
  timestamp = {Wed, 30 Jun 2021 16:14:10 +0200},
  biburl    = {https://dblp.org/rec/journals/corr/abs-2106-12732.bib},
  bibsource = {dblp computer science bibliography, https://dblp.org}
}

@inproceedings{10.1007/978-3-031-13185-1_7,
author = {Fischer, Marc and Sprecher, Christian and Dimitrov, Dimitar Iliev and Singh, Gagandeep and Vechev, Martin},
title = {Shared Certificates for Neural Network Verification},
year = {2022},
isbn = {978-3-031-13184-4},
publisher = {Springer-Verlag},
_address = {Berlin, Heidelberg},
_url = {https://doi.org/10.1007/978-3-031-13185-1_7},
doi = {10.1007/978-3-031-13185-1_7},
abstract = {Existing neural network verifiers compute a proof that each input is handled correctly under a given perturbation by propagating a symbolic abstraction of reachable values at each layer. This process is repeated from scratch independently for each input (e.g., image) and perturbation (e.g., rotation), leading to an expensive overall proof effort when handling an entire dataset. In this work, we introduce a new method for reducing this verification cost without losing precision based on a key insight that abstractions obtained at intermediate layers for different inputs and perturbations can overlap or contain each other. Leveraging our insight, we introduce the general concept of shared certificates, enabling proof effort reuse across multiple inputs to reduce overall verification costs. We perform an extensive experimental evaluation to demonstrate the effectiveness of shared certificates in reducing the verification cost on a range of datasets and attack specifications on image classifiers including the popular patch and geometric perturbations. We release our implementation at .},
booktitle = {Computer Aided Verification: 34th International Conference, CAV 2022},
pages = {127–148},
numpages = {22},
keywords = {Adversarial Robustness, Local Verification, Neural Network Verification},
location = {Haifa, Israel}
}

@article{balunovic2019certifying,
  title={Certifying geometric robustness of neural networks},
  author={Balunovic, Mislav and Baader, Maximilian and Singh, Gagandeep and Gehr, Timon and Vechev, Martin},
  journal={Advances in Neural Information Processing Systems},
  volume={32},
  year={2019}
}

@inproceedings{Botoeva:20,
 author = {Elena Botoeva and
Panagiotis Kouvaros and
Jan Kronqvist and
Alessio Lomuscio and
Ruth Misener},
 bibsource = {dblp computer science bibliography, https://dblp.org},
 booktitle = {Proc. of Advances in Artificial Intelligence (AAAI)},
 title = {Efficient Verification of ReLU-Based Neural Networks via Dependency
Analysis},
 year = {2020}
}

@misc{albarghouthi2021introductionneuralnetworkverification,
      title={Introduction to Neural Network Verification}, 
      author={Aws Albarghouthi},
      year={2021},
      eprint={2109.10317},
      archivePrefix={arXiv},
      primaryClass={cs.LG},
      url={https://arxiv.org/abs/2109.10317}, 
}

@article{serra2021scalingexactneuralnetwork,
  title={Scaling up exact neural network compression by ReLU stability},
  author={Serra, Thiago and Yu, Xin and Kumar, Abhinav and Ramalingam, Srikumar},
  journal={Advances in neural information processing systems},
  volume={34},
  pages={27081--27093},
  year={2021}
}

@article{singh2018fast,
	title={Fast and effective robustness certification},
	author={Singh, Gagandeep and Gehr, Timon and Mirman, Matthew and P{\"u}schel, Markus and Vechev, Martin},
	journal={Advances in Neural Information Processing Systems},
	volume={31},
	pages={10802--10813},
	year={2018}
}

@article{Lirpa:20,
  title={Automatic perturbation analysis for scalable certified robustness and beyond},
  author={Xu, Kaidi and Shi, Zhouxing and Zhang, Huan and Wang, Yihan and Chang, Kai-Wei and Huang, Minlie and Kailkhura, Bhavya and Lin, Xue and Hsieh, Cho-Jui},
  journal={Advances in Neural Information Processing Systems},
  volume={33},
  pages={1129--1141},
  year={2020}
}

@article{singh2019beyond,
  title={Beyond the single neuron convex barrier for neural network certification},
  author={Singh, Gagandeep and Ganvir, Rupanshu and P{\"u}schel, Markus and Vechev, Martin},
  journal={Advances in Neural Information Processing Systems},
  volume={32},
  year={2019}
}

@article{DBLP:conf/nips/SalmanY0HZ19,
  title={A convex relaxation barrier to tight robustness verification of neural networks},
  author={Salman, Hadi and Yang, Greg and Zhang, Huan and Hsieh, Cho-Jui and Zhang, Pengchuan},
  journal={Advances in Neural Information Processing Systems},
  volume={32},
  year={2019}
}

@inproceedings{zhang2018crown,
	title={Efficient neural network robustness certification with general activation functions},
	author={Zhang, Huan and Weng, Tsui-Wei and Chen, Pin-Yu and Hsieh, Cho-Jui and Daniel, Luca},
	booktitle={Advances in neural information processing systems},
	year={2018}
}

@article{singh2019abstract,
	title={An abstract domain for certifying neural networks},
	author={Singh, Gagandeep and Gehr, Timon and P{\"u}schel, Markus and Vechev, Martin},
	journal={Proceedings of the ACM on Programming Languages},
	volume={3},
	number={POPL},
	pages={1--30},
	year={2019},
	publisher={ACM New York, NY, USA}
}

@inproceedings{gehr2018ai2,
	title={Ai2: Safety and robustness certification of neural networks with abstract interpretation},
	author={Gehr, Timon and Mirman, Matthew and Drachsler-Cohen, Dana and Tsankov, Petar and Chaudhuri, Swarat and Vechev, Martin},
	booktitle={2018 IEEE Symposium on Security and Privacy (SP)},
	pages={3--18},
	year={2018},
	organization={IEEE}
}

@article{10.1145/3720417,
author = {Zhang, Guanqin and Zhang, Zhenya and Bandara, H.M.N. Dilum and Chen, Shiping and Zhao, Jianjun and Sui, Yulei},
title = {Efficient Incremental Verification of Neural Networks Guided by Counterexample Potentiality},
year = {2025},
issue_date = {April 2025},
_publisher = {Association for Computing Machinery},
_address = {New York, NY, USA},
volume = {9},
number = {OOPSLA1},
_url = {https://doi.org/10.1145/3720417},
doi = {10.1145/3720417},
abstract = {Incremental verification is an emerging neural network verification approach that aims to accelerate the verification of a neural network N* by reusing the existing verification result (called a template) of a similar neural network N. To date, the state-of-the-art incremental verification approach leverages the problem splitting history produced by branch and bound (BaB in verification of N, to select only a part of the sub-problems for verification of N*, thus more efficient than verifying N* from scratch. While this approach identifies whether each sub-problem should be re-assessed, it neglects the information of how necessary each sub-problem should be re-assessed, in the sense that the sub-problems that are more likely to contain counterexamples should be prioritized, in order to terminate the verification process as soon as a counterexample is detected. To bridge this gap, we first define a counterexample potentiality order over different sub-problems based on the template, and then we propose Olive, an incremental verification approach that explores the sub-problems of verifying N* orderly guided by counterexample potentiality. Specifically, Olive has two variants, including Oliveg, a greedy strategy that always prefers to exploit the sub-problems that are more likely to contain counterexamples, and Oliveb, a balanced strategy that also explores the sub-problems that are less likely, in case the template is not sufficiently precise. We experimentally evaluate the efficiency of Olive on 1445 verification problem instances derived from 15 neural networks spanning over two datasets MNIST and CIFAR-10. Our evaluation demonstrates significant performance advantages of Olive over state-of-the-art classic verification and incremental approaches. In particular, Olive shows evident superiority on the problem instances that contain counterexamples, and performs as well as Ivan on the certified problem instances.},
journal = {Proc. ACM Program. Lang.},
month = apr,
articleno = {83},
numpages = {28},
keywords = {branch and bound, counterexample potentiality, incremental verification, neural network verification}
}

@inproceedings{
ugare2024incremental,
title={Incremental Randomized Smoothing Certification},
author={Shubham Ugare and Tarun Suresh and Debangshu Banerjee and Gagandeep Singh and Sasa Misailovic},
booktitle={The Twelfth International Conference on Learning Representations},
year={2024},
url={https://openreview.net/forum?id=SdeAPV1irk}
}

@article{10.1145/3591299,
author = {Ugare, Shubham and Banerjee, Debangshu and Misailovic, Sasa and Singh, Gagandeep},
title = {Incremental Verification of Neural Networks},
year = {2023},
issue_date = {June 2023},
publisher = {Association for Computing Machinery},
address = {New York, NY, USA},
volume = {7},
number = {PLDI},
url = {https://doi.org/10.1145/3591299},
doi = {10.1145/3591299},
journal = {Proc. ACM Program. Lang.},
month = jun,
articleno = {185},
numpages = {26}
}

@inproceedings{SzegedyZSBEGF13,
  author    = {Christian Szegedy and
               Wojciech Zaremba and
               Ilya Sutskever and
               Joan Bruna and
               Dumitru Erhan and
               Ian J. Goodfellow and
               Rob Fergus},
  title     = {Intriguing properties of neural networks},
  booktitle = {2nd International Conference on Learning Representations, {ICLR}},
  year      = {2014},
}

@inproceedings{GoodfellowSS14,
  author    = {Ian J. Goodfellow and
               Jonathon Shlens and
               Christian Szegedy},
  title     = {Explaining and Harnessing Adversarial Examples},
  booktitle = {3rd International Conference on Learning Representations, {ICLR}},
  year      = {2015},
}

@inproceedings{Carlini017,
 author = {Nicholas Carlini and
David A. Wagner},
 bibsource = {dblp computer science bibliography, https://dblp.org},
 booktitle = {Symposium on Security and Privacy (S\&P)},
 title = {Towards Evaluating the Robustness of Neural Networks},
 year = {2017}
}

@inproceedings{eykholt2018robustphysicalworldattacksdeep,
  title={Robust physical-world attacks on deep learning visual classification},
  author={Eykholt, Kevin and Evtimov, Ivan and Fernandes, Earlence and Li, Bo and Rahmati, Amir and Xiao, Chaowei and Prakash, Atul and Kohno, Tadayoshi and Song, Dawn},
  booktitle={Proceedings of the IEEE conference on computer vision and pattern recognition},
  pages={1625--1634},
  year={2018}
}

@InProceedings{mirman2018differentiable,
  title = 	 {Differentiable Abstract Interpretation for Provably Robust Neural Networks},
  author =       {Mirman, Matthew and Gehr, Timon and Vechev, Martin},
  booktitle = 	 {Proceedings of the 35th International Conference on Machine Learning},
  pages = 	 {3578--3586},
  year = 	 {2018},
  _editor = 	 {Dy, Jennifer and Krause, Andreas},
  volume = 	 {80},
  series = 	 {Proceedings of Machine Learning Research},
  month = 	 {10--15 Jul},
  publisher =    {PMLR},
  pdf = 	 {http://proceedings.mlr.press/v80/mirman18b/mirman18b.pdf},
  url = 	 {https://proceedings.mlr.press/v80/mirman18b.html},
  abstract = 	 {We introduce a scalable method for training robust neural networks based on abstract interpretation. We present several abstract transformers which balance efficiency with precision and show these can be used to train large neural networks that are certifiably robust to adversarial perturbations.}
}

@inproceedings{wong2017provable,
  title={Provable defenses against adversarial examples via the convex outer adversarial polytope},
  author={Wong, Eric and Kolter, Zico},
  booktitle={International conference on machine learning},
  pages={5286--5295},
  year={2018},
  organization={PMLR}
}

@inproceedings{madry:17,
  title={Towards deep learning models resistant to adversarial attacks},
  author={Madry, Aleksander and Makelov, Aleksandar and Schmidt, Ludwig and Tsipras, Dimitris and Vladu, Adrian},
  booktitle={Proc. International Conference on Learning Representations (ICLR)},
  year={2018}
}

@article{taps,
  title={Connecting certified and adversarial training},
  author={Mao, Yuhao and M{\"u}ller, Mark and Fischer, Marc and Vechev, Martin},
  journal={Advances in Neural Information Processing Systems},
  volume={36},
  pages={73422--73440},
  year={2023}
}

@inproceedings{
    sabr,
    title={Certified Training: Small Boxes are All You Need},
    author={Mark Niklas M{\"{u}}ller and Franziska Eckert and Marc Fischer and Martin Vechev},
    booktitle={International Conference on Learning Representations},
    year={2023},
    url={https://openreview.net/forum?id=7oFuxtJtUMH}
}

@inproceedings{wu2024marabou20versatileformal,
  title={Marabou 2.0: a versatile formal analyzer of neural networks},
  author={Wu, Haoze and Isac, Omri and Zelji{\'c}, Aleksandar and Tagomori, Teruhiro and Daggitt, Matthew and Kokke, Wen and Refaeli, Idan and Amir, Guy and Julian, Kyle and Bassan, Shahaf and others},
  booktitle={International Conference on Computer Aided Verification},
  pages={249--264},
  year={2024},
  organization={Springer}
}

@article{shapira2026tightrobustnesscertificationconvex,
  author    = {Shapira, Yuval and Drachsler-Cohen, Dana},
  title     = {Tight Robustness Certification Through the Convex Hull of {$\ell_0$} Attacks},
  journal   = {Proceedings of the AAAI Conference on Artificial Intelligence},
  volume    = {40},
  number    = {44},
  pages     = {37913--37922},
  year      = {2026},
  doi       = {10.1609/aaai.v40i44.41128}
}

@inproceedings{li2023sok,
    title={{SoK}: Certified Robustness for Deep Neural Networks},
    author={Linyi Li and Tao Xie and Bo Li},
    booktitle={44th {IEEE} Symposium on Security and Privacy, {SP} 2023, San Francisco, CA, USA, 22-26 May 2023},
    publisher={IEEE},
    year={2023}
}

@article{safetytrustabsint,
author = {Singh, Gagandeep and Laurel, Jacob and Misailovic, Sasa and Banerjee, Debangshu and Singh, Avaljot and Xu, Changming and Ugare, Shubham and Zhang, Huan},
title = {Safety and Trust in Artificial Intelligence with Abstract Interpretation},
year = {2025},
issue_date = {Jun 2025},
publisher = {Now Publishers Inc.},
address = {Hanover, MA, USA},
volume = {8},
number = {3–4},
issn = {2325-1107},
url = {https://doi.org/10.1561/2500000062},
doi = {10.1561/2500000062},
journal = {Found. Trends Program. Lang.},
month = jun,
pages = {250–408},
numpages = {162}
}

@article{SuVS19,
  author       = {Jiawei Su and
                  Danilo Vasconcellos Vargas and
                  Kouichi Sakurai},
  title        = {One Pixel Attack for Fooling Deep Neural Networks},
  journal      = {{IEEE} Trans. Evol. Comput.},
  volume       = {23},
  number       = {5},
  year         = {2019}}

@inproceedings{ChiangNAZSG20,
 author = {Ping{-}Yeh Chiang and
Renkun Ni and
Ahmed Abdelkader and
Chen Zhu and
Christoph Studer and
Tom Goldstein},
 bibsource = {dblp computer science bibliography, https://dblp.org},
 booktitle = {Proc. of International Conf. on Learning Representations (ICLR)},
 title = {Certified Defenses for Adversarial Patches},
 year = {2020}
}

@article{jiang2025universalcertifiedrobustnessmultinorm,
  author       = {Enyi Jiang and
                  David Shu Cheung and
                  Gagandeep Singh},
  title        = {Towards Generalized Certified Robustness with Multi-Norm Training},
  journal      = {Trans. Mach. Learn. Res.},
  volume       = {2026},
  year         = {2026},
  url          = {https://openreview.net/forum?id=U5U7pazr6X},
  bibsource    = {dblp computer science bibliography, https://dblp.org}
}

@inproceedings{modas2019sparsefool,
  title={Sparsefool: a few pixels make a big difference},
  author={Modas, Apostolos and Moosavi-Dezfooli, Seyed-Mohsen and Frossard, Pascal},
  booktitle={Proceedings of the IEEE Conference on Computer Vision and Pattern Recognition},
  pages={9087--9096},
  year={2019}
}

@article{Shapira23,
author = {Shapira, Yuval and Avneri, Eran and Drachsler-Cohen, Dana},
title = {Deep Learning Robustness Verification for Few-Pixel Attacks},
journal      = {Proc. {ACM} Program. Lang.},
  volume       = {7},
  number       = {{OOPSLA1}},
  year         = {2023},

}

@inproceedings{CroceASF022,
  author       = {Francesco Croce and
                  Maksym Andriushchenko and
                  Naman D. Singh and
                  Nicolas Flammarion and
                  Matthias Hein},
  title        = {{Sparse-RS}: {A} Versatile Framework for Query-Efficient Sparse Black-Box
                  Adversarial Attacks},
  booktitle    = {Thirty-Sixth {AAAI} Conference on Artificial Intelligence, {AAAI}},
  _publisher    = {{AAAI} Press},
  year         = {2022}}

@inproceedings{shapira2024boostingfewpixelrobustnessverification,
  title={Boosting few-pixel robustness verification via covering verification designs},
  author={Shapira, Yuval and Wiesel, Naor and Shabelman, Shahar and Drachsler-Cohen, Dana},
  booktitle={International Conference on Computer Aided Verification},
  pages={377--400},
  year={2024},
  organization={Springer}
}

@inproceedings{10.1145/2465449.2465456,
author = {Johnson, Kenneth and Calinescu, Radu and Kikuchi, Shinji},
title = {An Incremental Verification Framework for Component-Based Software Systems},
year = {2013},
isbn = {9781450321228},
_publisher = {Association for Computing Machinery},
_address = {New York, NY, USA},
_url = {https://doi.org/10.1145/2465449.2465456},
doi = {10.1145/2465449.2465456},
booktitle = {Proceedings of the ACM {SIGSOFT} Symposium on Component-Based Software Engineering},
pages = {33–42},
numpages = {10},
keywords = {domain-specific languages, incremental verification, probabilistic assume-guarantee verification},
location = {Vancouver, British Columbia, Canada},
series = {CBSE '13}
}

@inproceedings{Lachnech-etal:TACAS01,
  title={Incremental verification by abstraction},
  author={Lakhnech, Yassine and Bensalem, Saddek and Berezin, Sergey and Owre, Sam},
  booktitle={International Conference on Tools and Algorithms for the Construction and Analysis of Systems},
  pages={98--112},
  year={2001},
  organization={Springer}
}

@inproceedings{DBLP:conf/lics/OHearn18,
  author    = {Peter W. O'Hearn},
  editor    = {Anuj Dawar and
               Erich Gr{\"{a}}del},
  title     = {Continuous Reasoning: Scaling the impact of formal methods},
  booktitle = {Proceedings of the 33rd Annual {ACM/IEEE} Symposium on Logic in Computer
               Science, {LICS} 2018},
  _pages     = {13--25},
  _publisher = {{ACM}},
  year      = {2018},
  url       = {https://doi.org/10.1145/3209108.3209109},
  doi       = {10.1145/3209108.3209109},
  timestamp = {Wed, 21 Nov 2018 12:44:18 +0100},
  biburl    = {https://dblp.org/rec/conf/lics/OHearn18.bib},
  bibsource = {dblp computer science bibliography, https://dblp.org}
}

@inproceedings{DBLP:conf/pldi/0002CS21,
  author    = {Benno Stein and
               Bor{-}Yuh Evan Chang and
               Manu Sridharan},
  editor    = {Stephen N. Freund and
               Eran Yahav},
  title     = {Demanded abstract interpretation},
  booktitle = {{PLDI} '21: 42nd {ACM} {SIGPLAN} International Conference on Programming
               Language Design and Implementation, Virtual Event, Canada, June 20-25,
               2021},
  pages     = {282--295},
  publisher = {{ACM}},
  year      = {2021},
  url       = {https://doi.org/10.1145/3453483.3454044},
  doi       = {10.1145/3453483.3454044},
  bibsource = {dblp computer science bibliography, https://dblp.org}
}

@inproceedings{10.1145/2491411.2491429,
author = {Beyer, Dirk and L\"{o}we, Stefan and Novikov, Evgeny and Stahlbauer, Andreas and Wendler, Philipp},
title = {Precision Reuse for Efficient Regression Verification},
year = {2013},
isbn = {9781450322379},
_publisher = {Association for Computing Machinery},
_address = {New York, NY, USA},
_url = {https://doi.org/10.1145/2491411.2491429},
doi = {10.1145/2491411.2491429},
_abstract = {},
booktitle = {Proceedings of the 2013 9th Joint Meeting on Foundations of Software Engineering},
pages = {389–399},
numpages = {11},
keywords = {Regression Checking, Formal Verification},
location = {Saint Petersburg, Russia},
series = {ESEC/FSE 2013}
}

@inproceedings{yang2023provable,
  author    = {Yang, Rem and Laurel, Jacob and Misailovic, Sasa and Singh, Gagandeep},
  title     = {Provable Defense Against Geometric Transformations},
  booktitle = {The Eleventh International Conference on Learning Representations},
  year      = {2023},
  url       = {https://openreview.net/forum?id=ThXqBsRI-cY}
}

@INPROCEEDINGS{5306334,
  author={Yang, Guowei and Dwyer, Matthew B. and Rothermel, Gregg},
  booktitle={2009 IEEE International Conference on Software Maintenance}, 
  title={Regression model checking}, 
  year={2009},
  volume={},
  number={},
  pages={115-124},
  doi={10.1109/ICSM.2009.5306334}}

@inproceedings{10.1145/2393596.2393665,
author = {Visser, Willem and Geldenhuys, Jaco and Dwyer, Matthew B.},
title = {Green: Reducing, Reusing and Recycling Constraints in Program Analysis},
year = {2012},
isbn = {9781450316149},
_publisher = {Association for Computing Machinery},
_address = {New York, NY, USA},
_url = {https://doi.org/10.1145/2393596.2393665},
doi = {10.1145/2393596.2393665},
_abstract = {},
booktitle = {Proceedings of the ACM SIGSOFT 20th International Symposium on the Foundations of Software Engineering},
articleno = {58},
numpages = {11},
keywords = {symbolic execution, constraint solving, NoSQL, path feasibility},
location = {Cary, North Carolina},
series = {FSE '12}
}

@inproceedings{Cousot:77,
 author = {Cousot, Patrick and Cousot, Radhia},
 booktitle = {Proc. of Principles of Programming Languages (POPL)},
 title = {Abstract Interpretation: A Unified Lattice Model for Static Analysis of Programs by Construction or Approximation of Fixpoints},
 year = {1977}
}

@article{kotyan2022adversarial,
  title={Adversarial robustness assessment: Why in evaluation both $L_0$ and $L_\infty$ attacks are necessary},
  author={Kotyan, Shashank and Vargas, Danilo Vasconcellos},
  journal={PloS one},
  volume={17},
  number={4},
  pages={e0265723},
  year={2022},
  publisher={Public Library of Science San Francisco, CA USA}
}

\clearpage
\appendix
\renewcommand{\thesection}{Appendix \arabic{section}}

\section{Joint Stability for Other Perturbations}
\label{sec:limit-study-appendix}
\begin{figure}[t!]
  \centering
  \begin{subfigure}[b]{\textwidth}
    \centering
    \includegraphics[width=\textwidth]{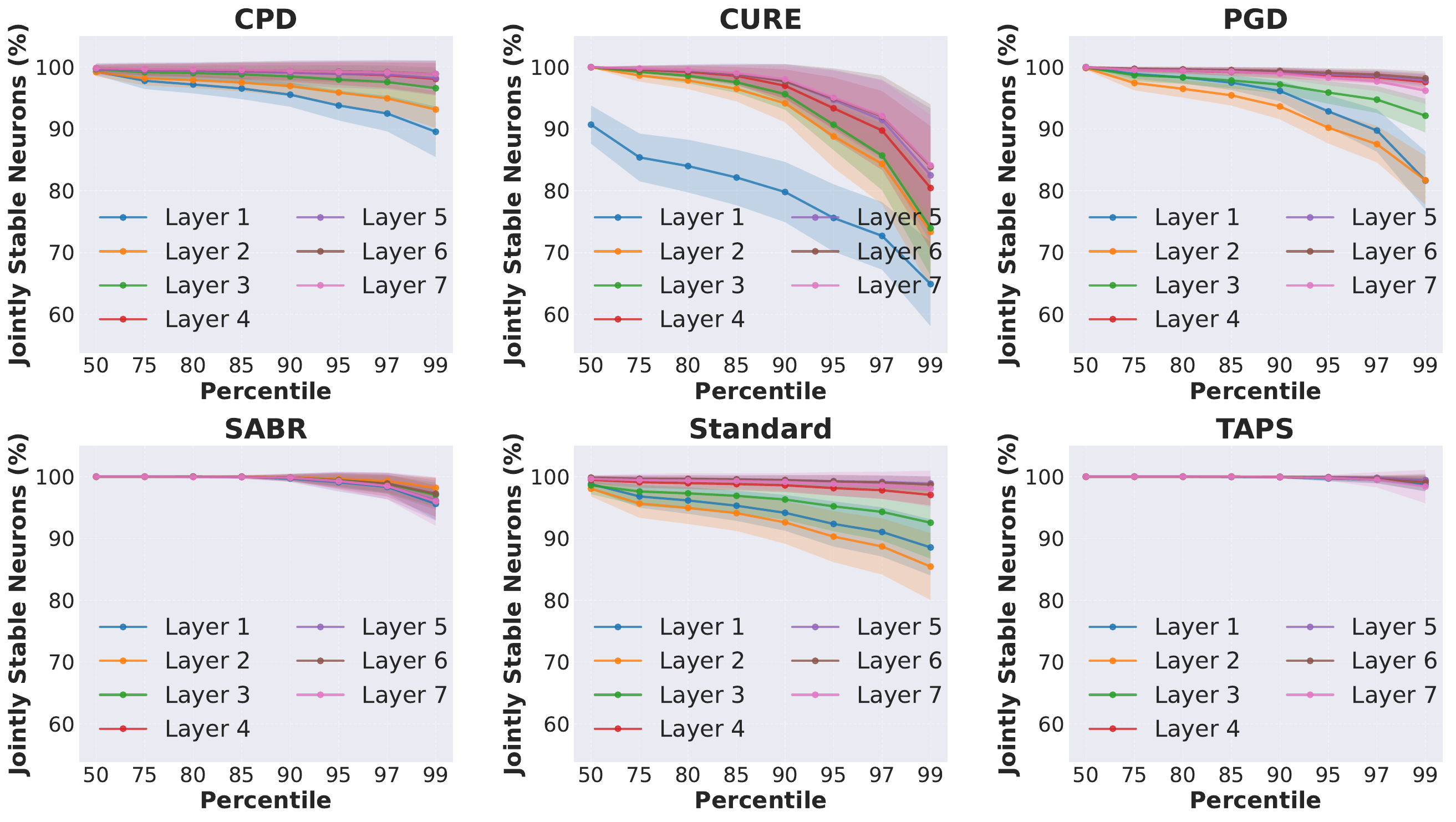}
    \caption{Jointly stable neurons}
  \end{subfigure}
  \vspace{0.5em}
  \begin{subfigure}[b]{\textwidth}
    \centering
    \includegraphics[width=\textwidth]{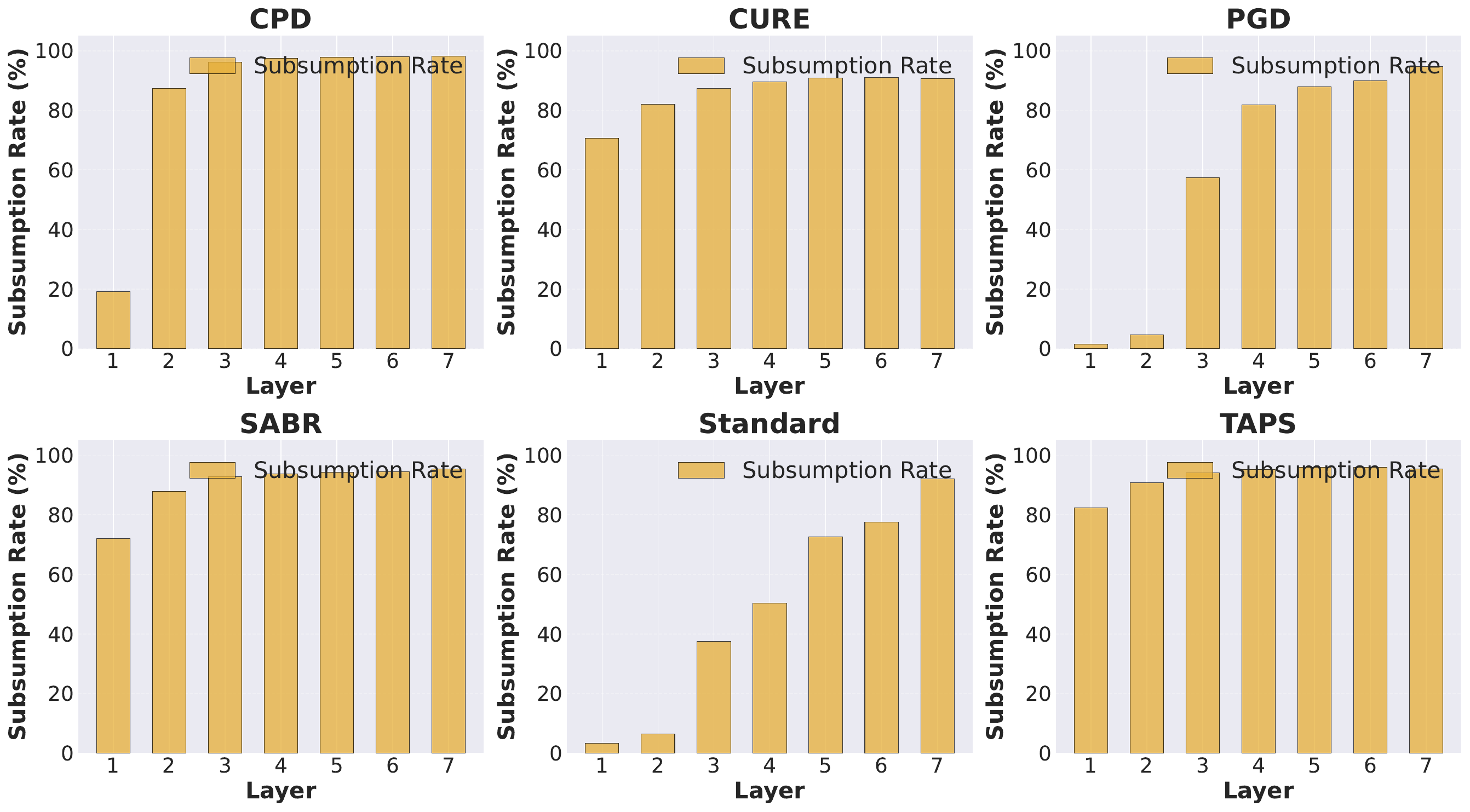}
    \caption{Template subsumption rates}
  \end{subfigure}
  \caption{Jointly stable neurons and template subsumption rates under $2 \times 2$ patch perturbations in MNIST networks.}
  \label{fig:mnist_patch}
\end{figure}

\begin{figure}[t!]
  \centering
  \begin{subfigure}[b]{\textwidth}
    \centering
    \includegraphics[width=\textwidth]{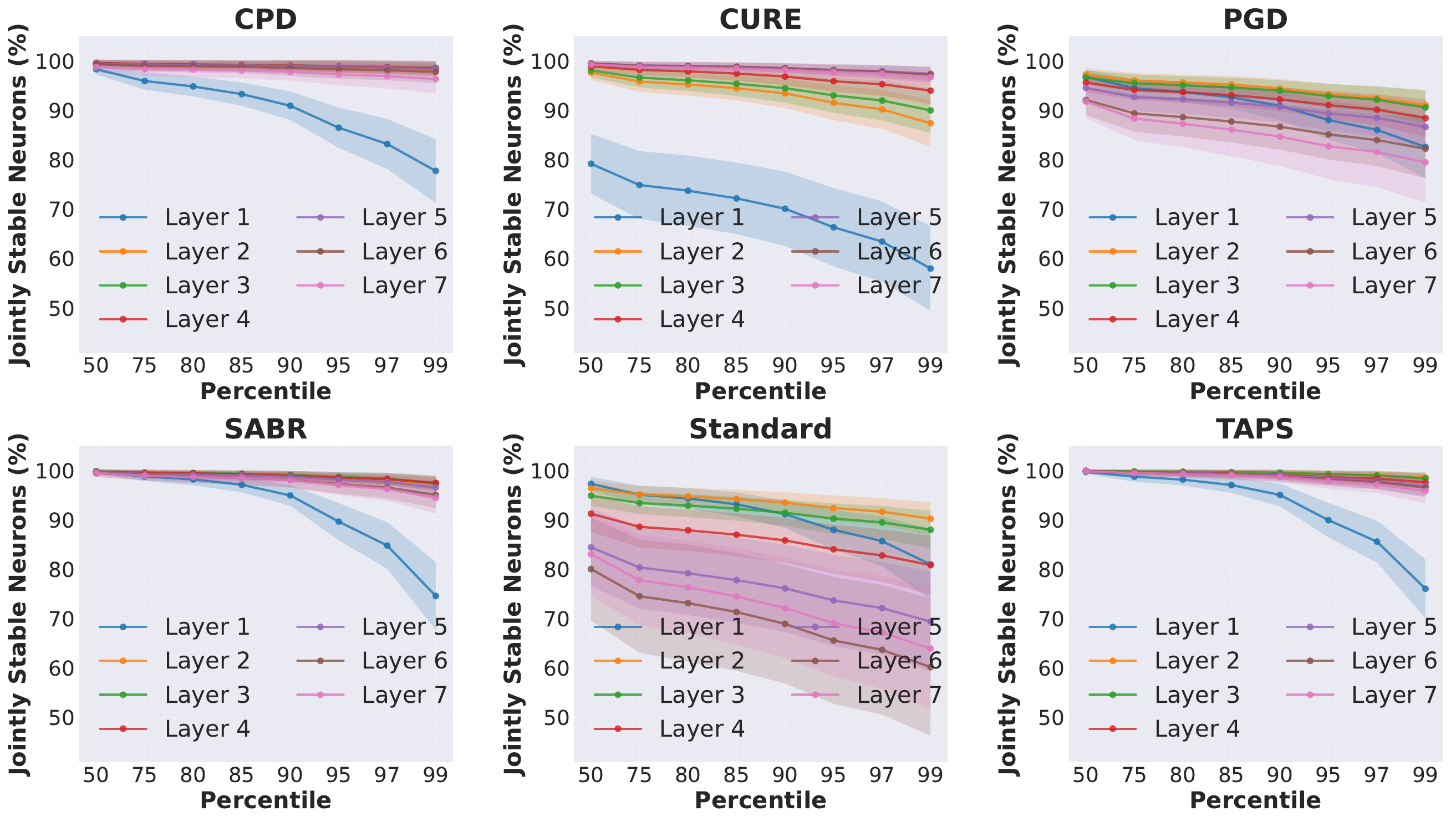}
    \caption{Jointly stable neurons}
  \end{subfigure}
  \vspace{0.5em}
  \begin{subfigure}[b]{\textwidth}
    \centering
    \includegraphics[width=\textwidth]{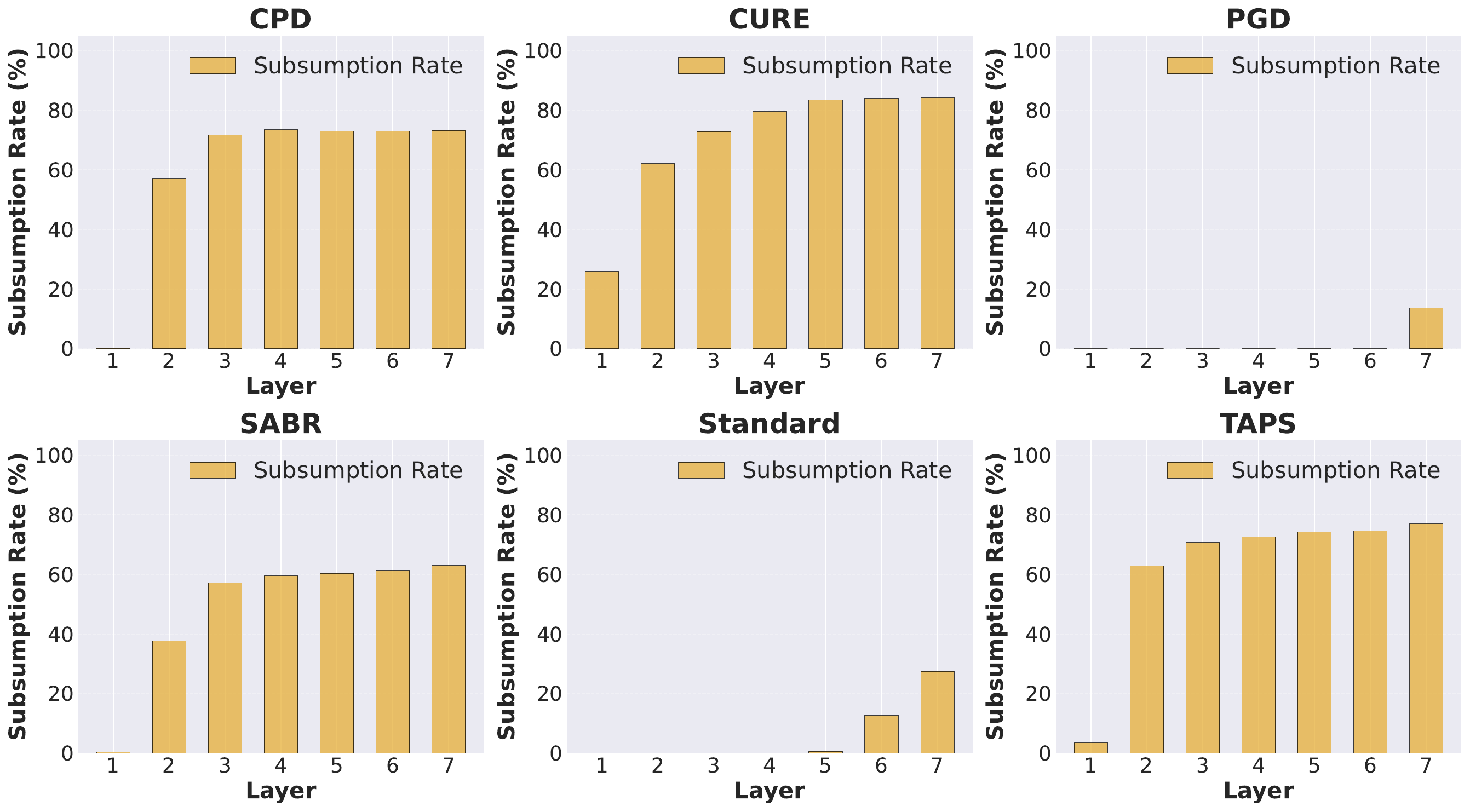}
    \caption{Template subsumption rates}
  \end{subfigure}
  \caption{Jointly stable neurons and template subsumption rates under $2 \times 2$ patch perturbations in CIFAR-10 networks.}
  \label{fig:cifar_patch}
\end{figure}

\begin{figure}[t!]
  \centering
  \begin{subfigure}[b]{\textwidth}
    \centering
    \includegraphics[width=\textwidth]{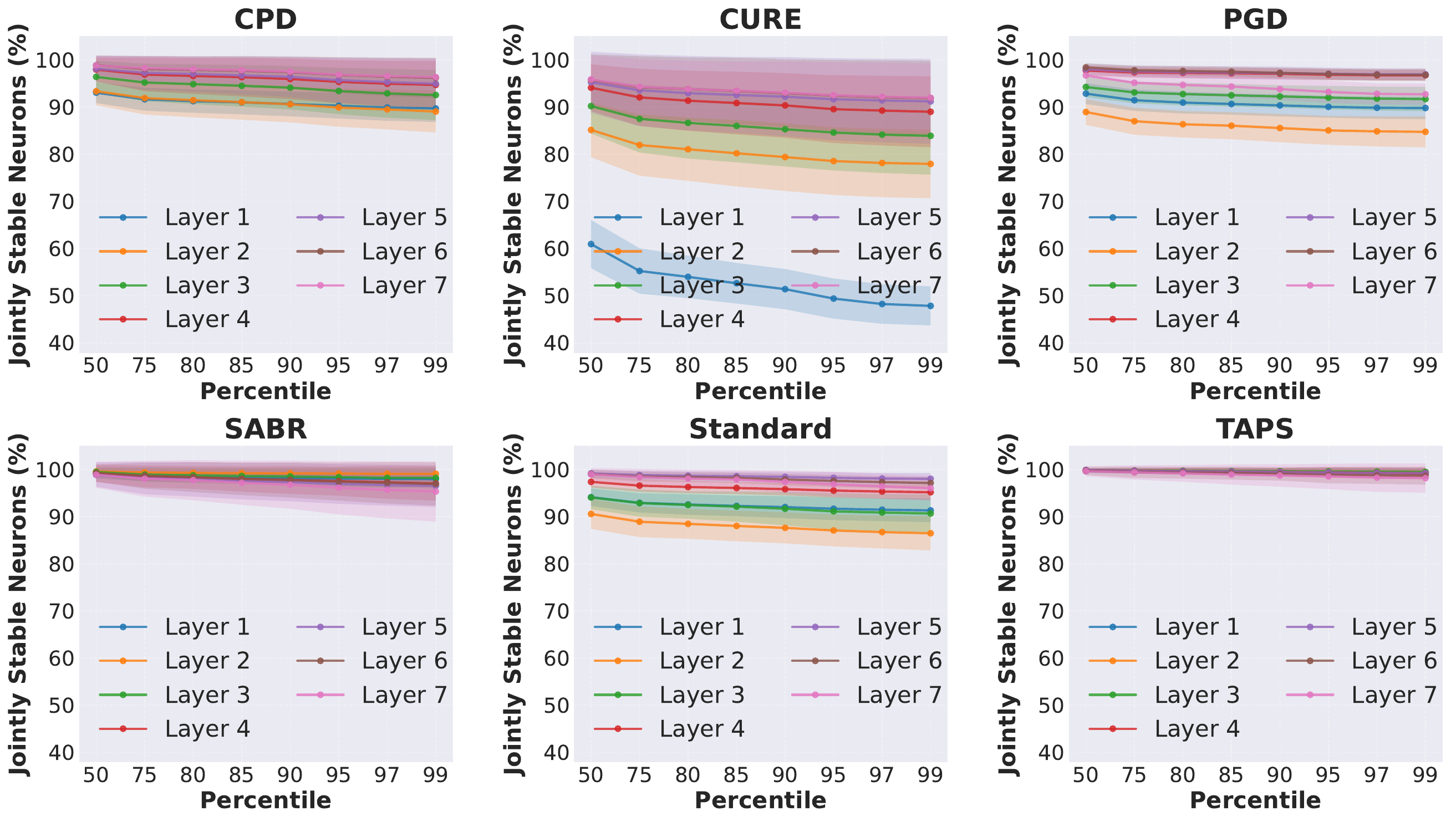}
    \caption{Jointly stable neurons}
  \end{subfigure}
  \vspace{0.5em}
  \begin{subfigure}[b]{\textwidth}
    \centering
    \includegraphics[width=\textwidth]{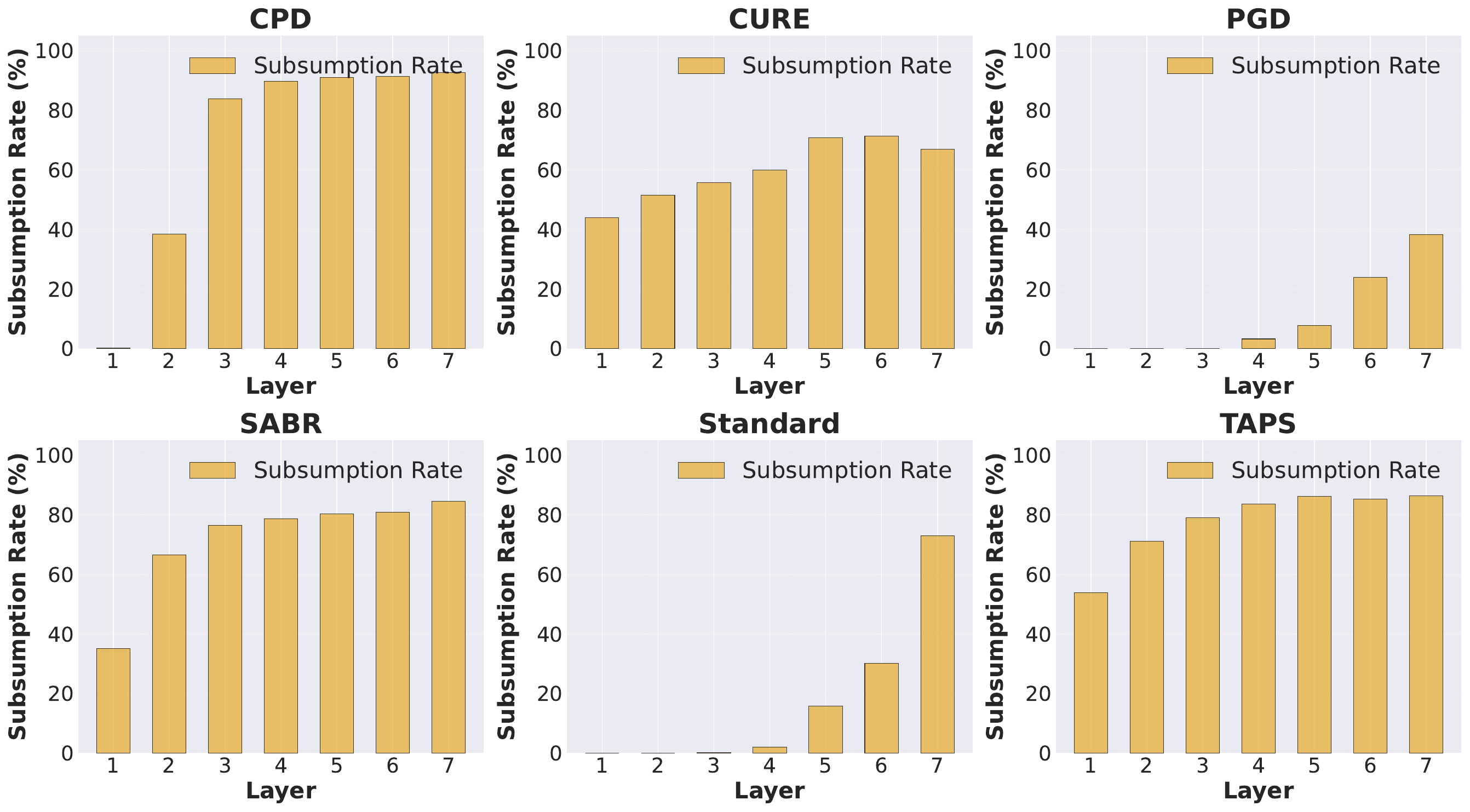}
    \caption{Template subsumption rates}
  \end{subfigure}
  \caption{Jointly stable neurons and template subsumption rates under geometric perturbations (4 splits) in MNIST networks.}
  \label{fig:mnist_g4}
\end{figure}

\begin{figure}[t!]
  \centering
  \begin{subfigure}[b]{\textwidth}
    \centering
    \includegraphics[width=\textwidth]{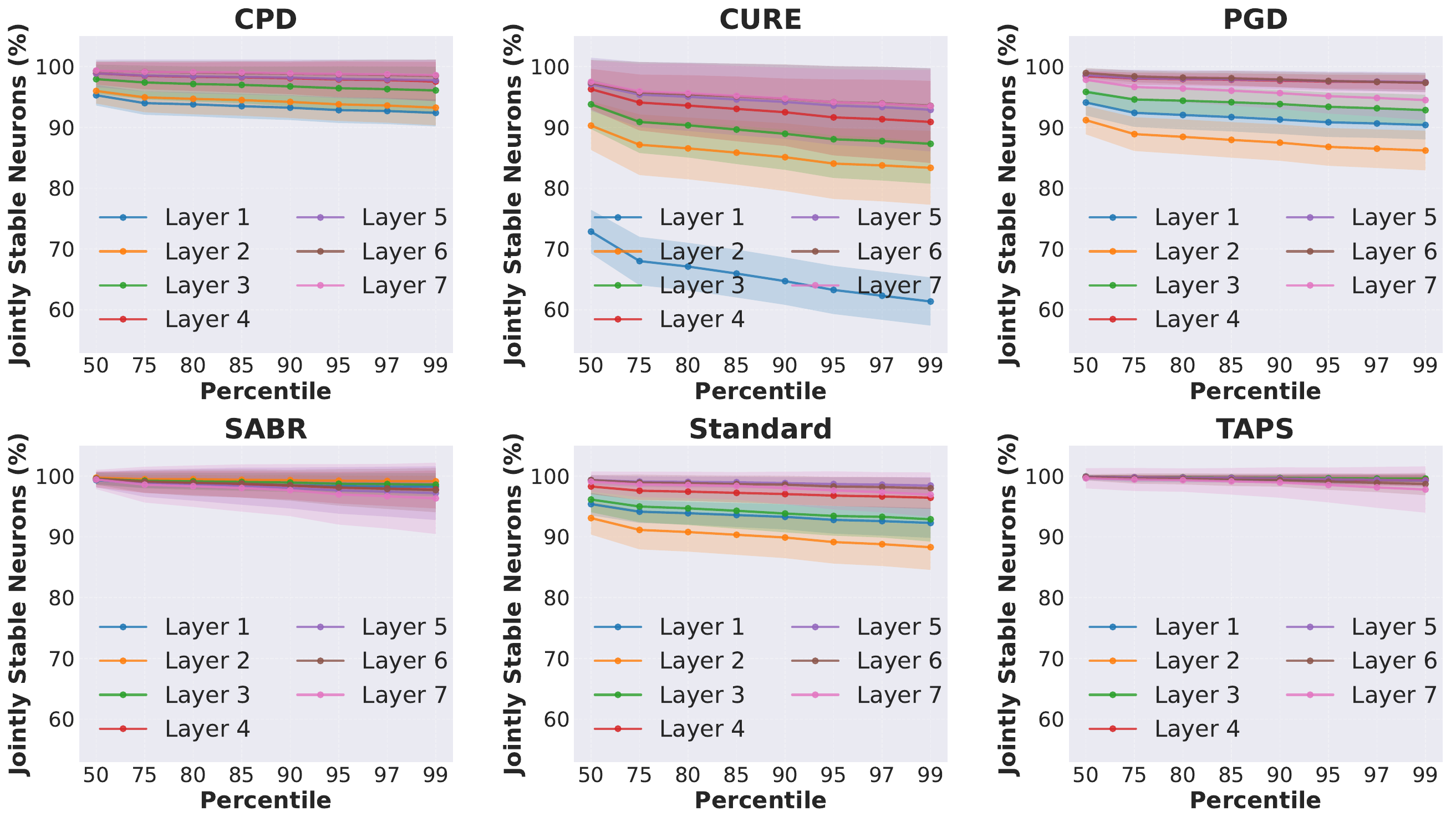}
    \caption{Jointly stable neurons}
  \end{subfigure}
  \vspace{0.5em}
  \begin{subfigure}[b]{\textwidth}
    \centering
    \includegraphics[width=\textwidth]{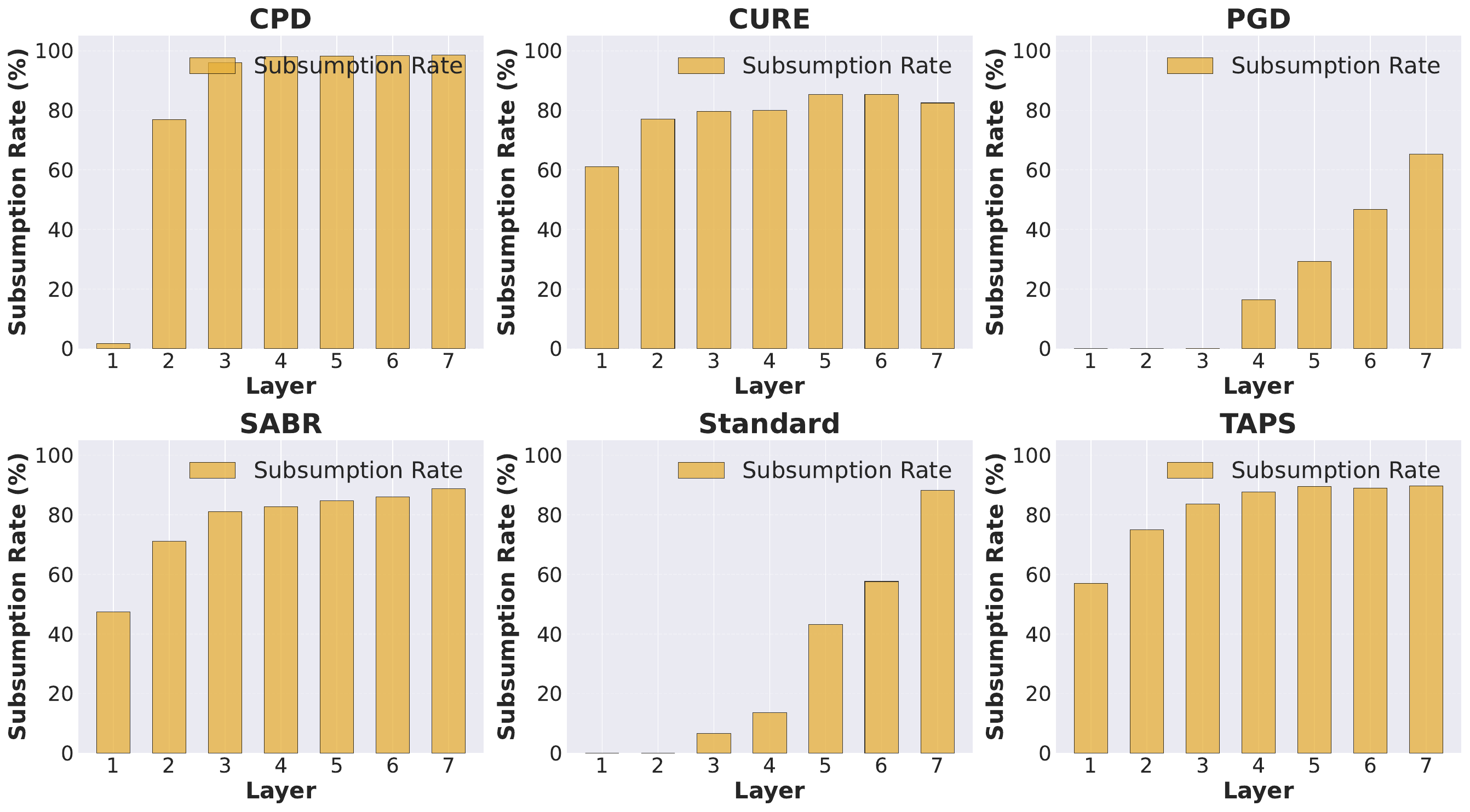}
    \caption{Template subsumption rates}
  \end{subfigure}
  \caption{Jointly stable neurons and template subsumption rates under geometric perturbations (6 splits) in MNIST networks.}
  \label{fig:mnist_g6}
\end{figure}

\begin{figure}[t!]
  \centering
  \begin{subfigure}[b]{\textwidth}
    \centering
    \includegraphics[width=\textwidth]{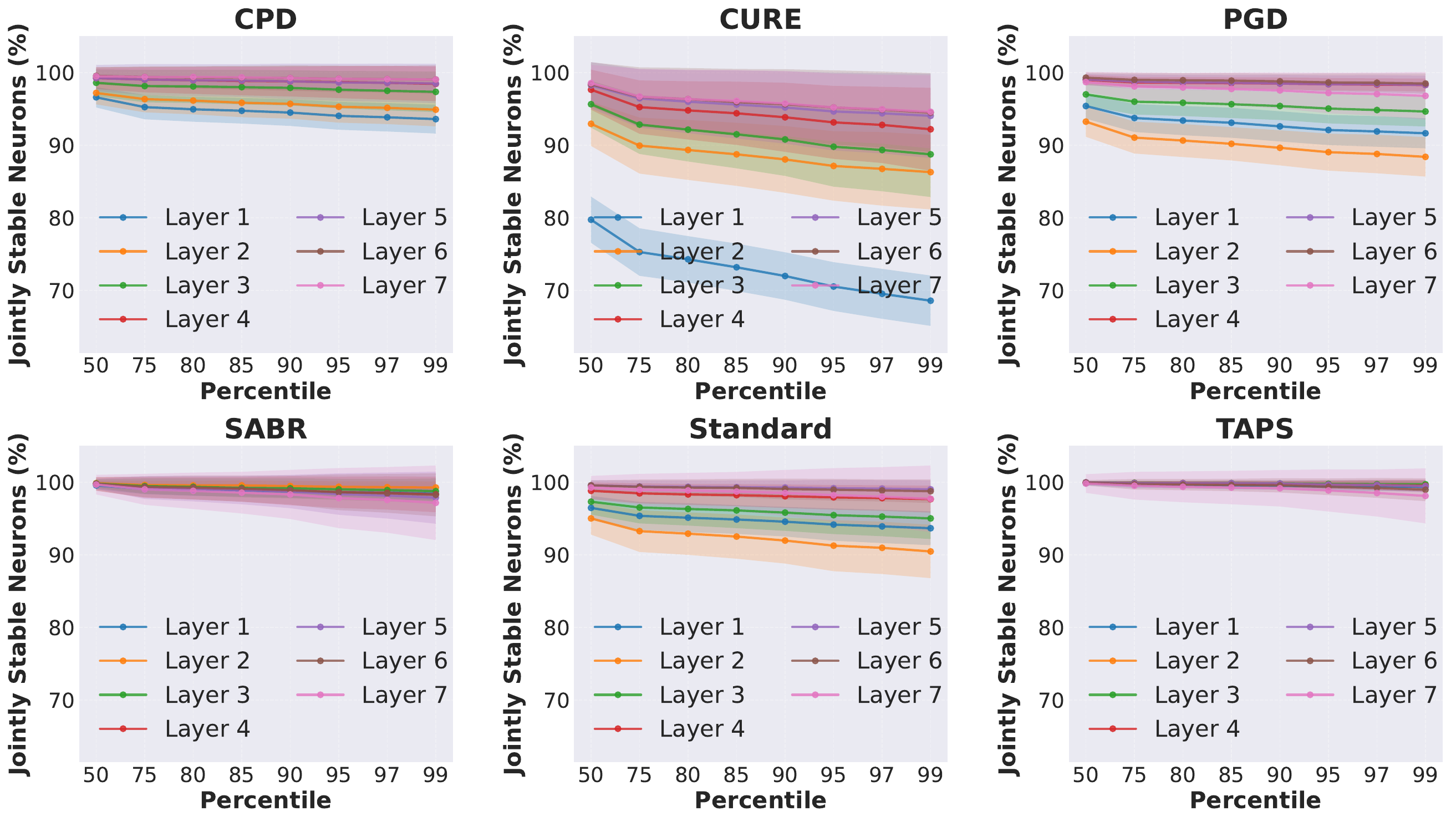}
    \caption{Jointly stable neurons}
  \end{subfigure}
  \vspace{0.5em}
  \begin{subfigure}[b]{\textwidth}
    \centering
    \includegraphics[width=\textwidth]{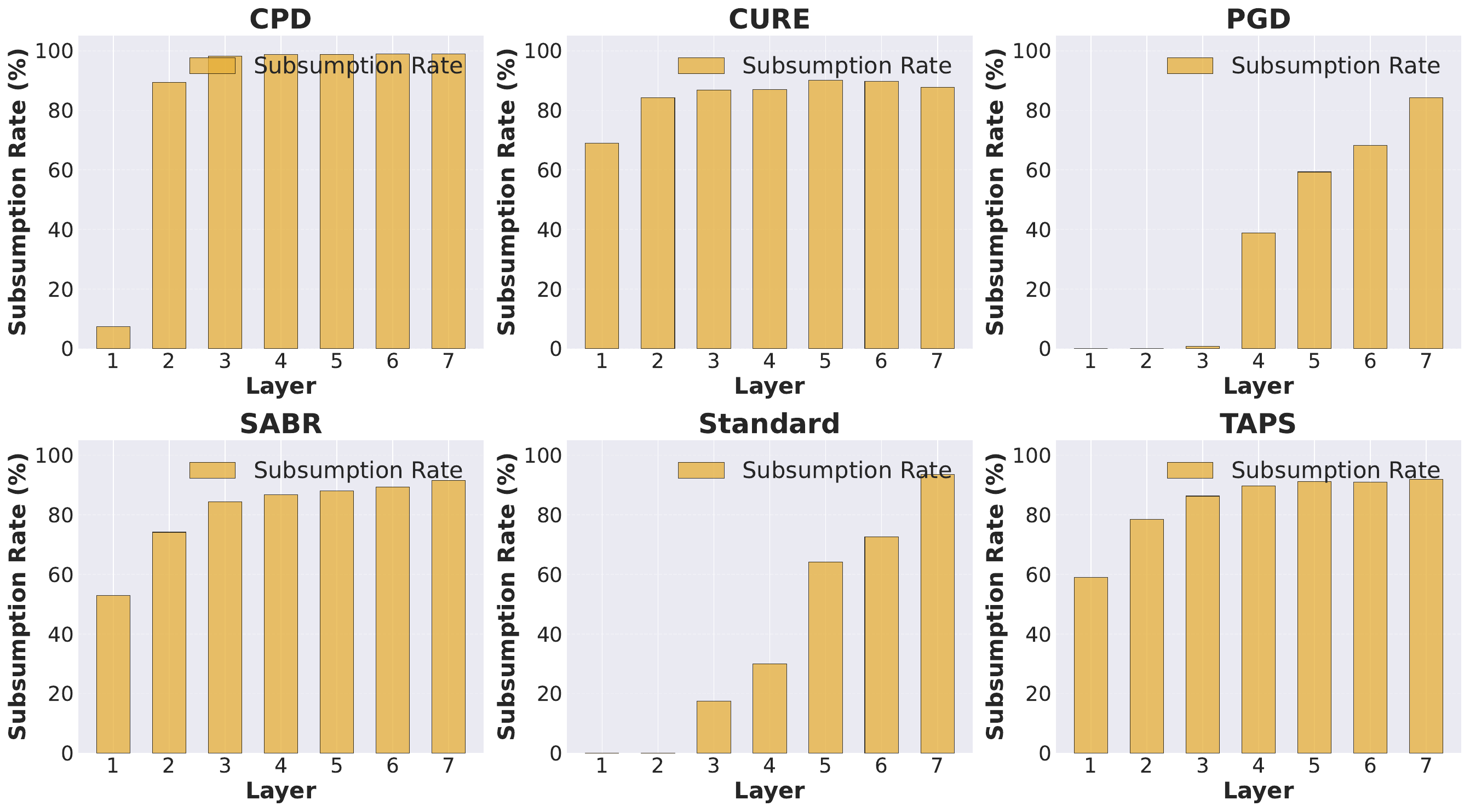}
    \caption{Template subsumption rates}
  \end{subfigure}
  \caption{Jointly stable neurons and template subsumption rates under geometric perturbations (8 splits) in MNIST networks.}
  \label{fig:mnist_g8}
\end{figure}

\begin{figure}[t!]
  \centering
  \begin{subfigure}[b]{\textwidth}
    \centering
    \includegraphics[width=\textwidth]{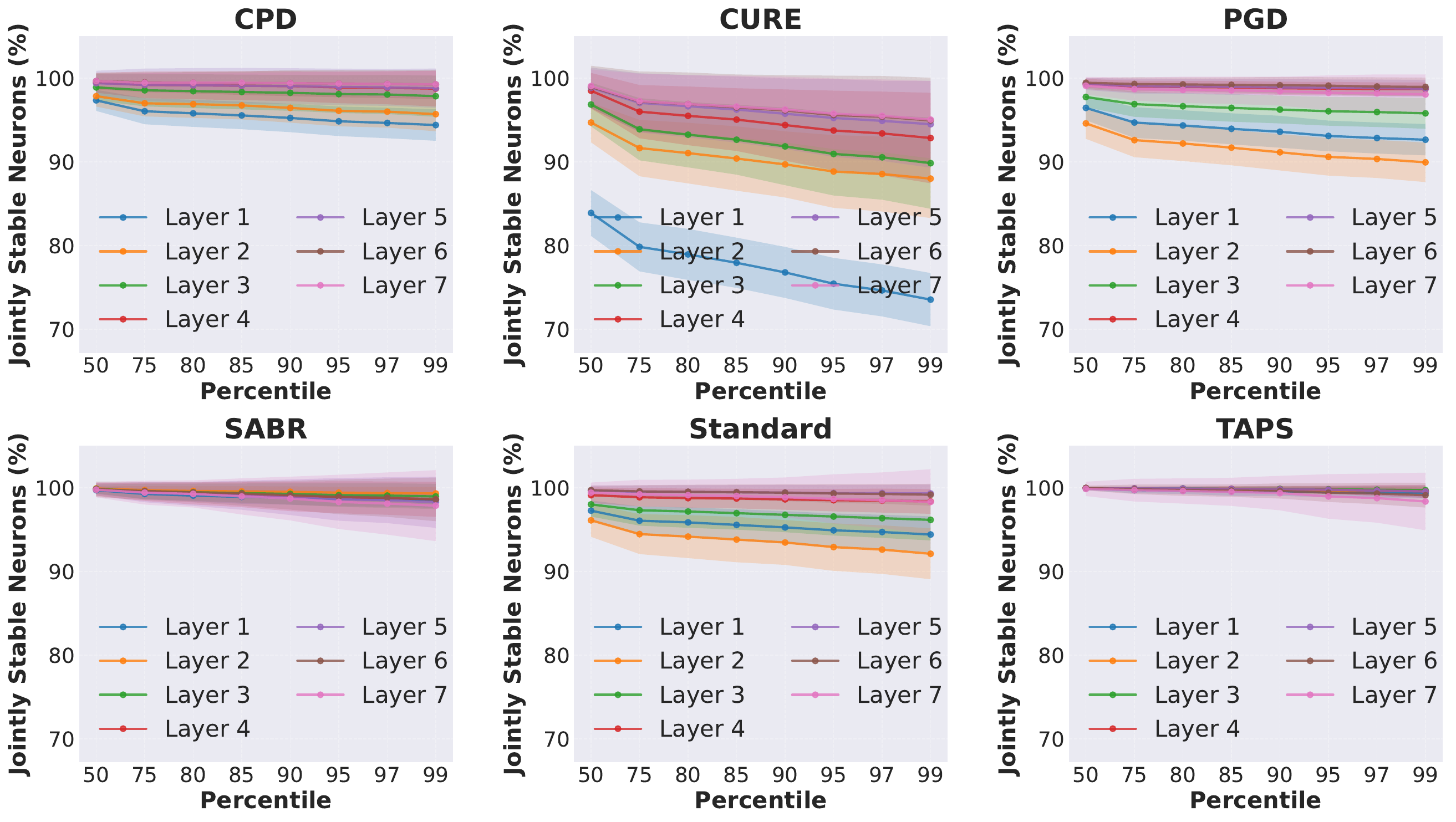}
    \caption{Jointly stable neurons}
  \end{subfigure}
  \vspace{0.5em}
  \begin{subfigure}[b]{\textwidth}
    \centering
    \includegraphics[width=\textwidth]{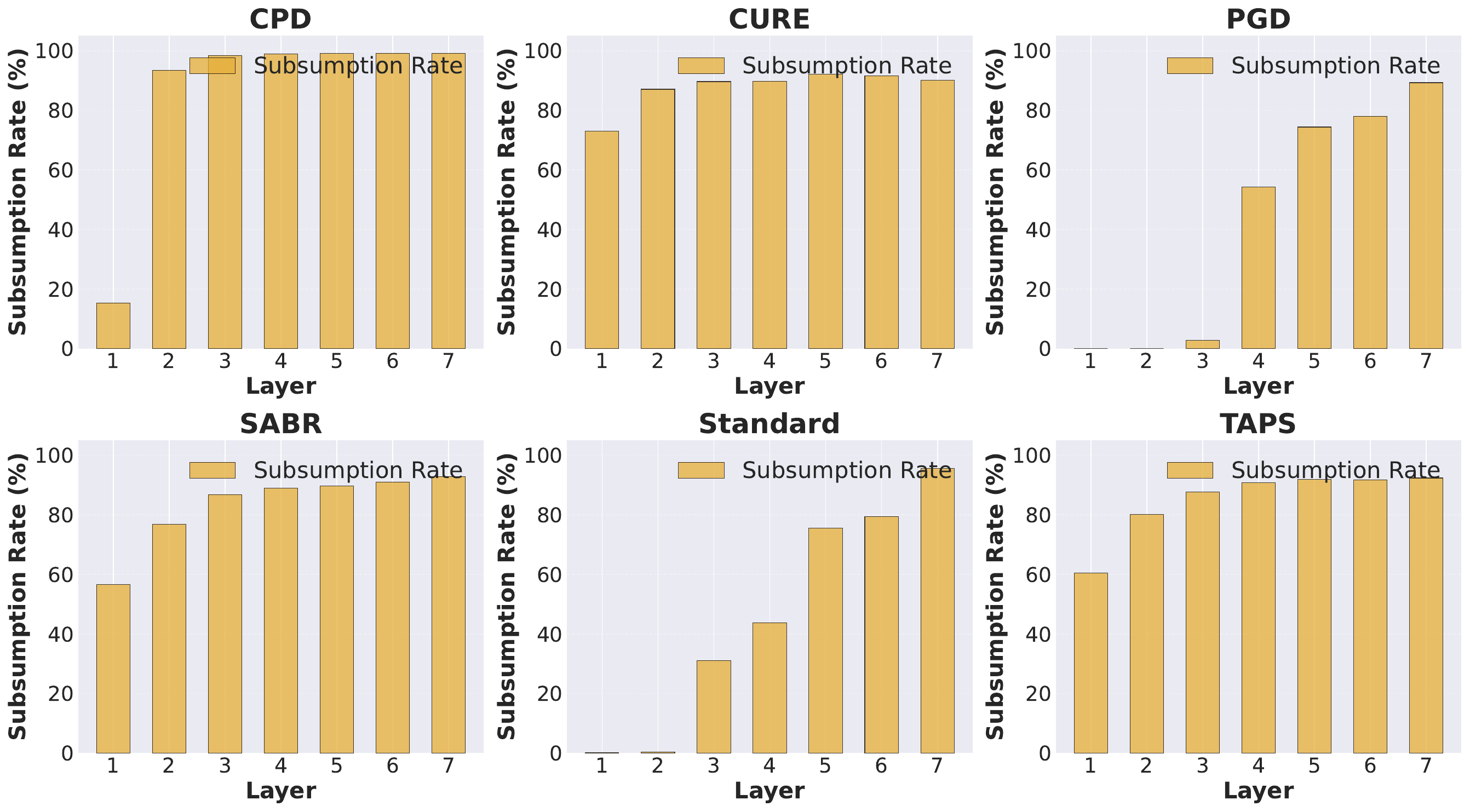}
    \caption{Template subsumption rates}
  \end{subfigure}
  \caption{Jointly stable neurons and template subsumption rates under geometric perturbations (10 splits) in MNIST networks.}
  \label{fig:mnist_g10}
\end{figure}

\Cref{sec:limit-study} presented trends in joint stability and subsumption rates for $L_0$ perturbations.  We observed similar joint stability and template subsumption trends across patch and geometric perturbations as well.
In~\cref{fig:mnist_patch,fig:cifar_patch}, we plot the jointly stable neurons and template subsumption rates across different $2 \times 2$ patch perturbation scenarios in MNIST and CIFAR-10 networks.
And in~\cref{fig:mnist_g4,fig:mnist_g6,fig:mnist_g8,fig:mnist_g10}, we plot the jointly stable neurons and template subsumption rates across geometric perturbations for MNIST networks.
Here, $\pm 2^\circ$ rotation, $\pm 10\%$ contrast and $\pm 1\%$ brightness changes are split into $r =$ 4, 6, 8, and 10 perturbations, respectively resulting in 64, 216, 512 and 1000 queries per image~\cite{10.1007/978-3-031-13185-1_7,balunovic2019certifying}.
The common trends of earlier layers having lower joint stability and subsumption rates, and later layers having higher joint stability and subsumption rates hold in these perturbation scenarios as well.
The interesting cases observed in~\cref{sec:limit-study}; e.g., CIFAR-10 Standard and PGD networks, with earlier layers having higher joint stability but still very low subsumption
rates in~\cref{fig:limit-study-cifar} for $L_0$ perturbations are also observed in~\cref{fig:cifar_patch} for patch perturbations. Similarly, trends reported for SABR, TAPS, and CURE MNIST networks in~\cref{fig:limit-study-mnist} remain identical in
the patch and geometric perturbation scenarios under~\cref{fig:mnist_patch,fig:mnist_g4,fig:mnist_g6,fig:mnist_g8,fig:mnist_g10}.

\section{Sampling Sizes for Subsumption Estimation}
\label{sec:implementation-appendix}

\begin{figure}[t!]
  \centering
  \begin{subfigure}[b]{\textwidth}
    \centering
    \includegraphics[width=\textwidth]{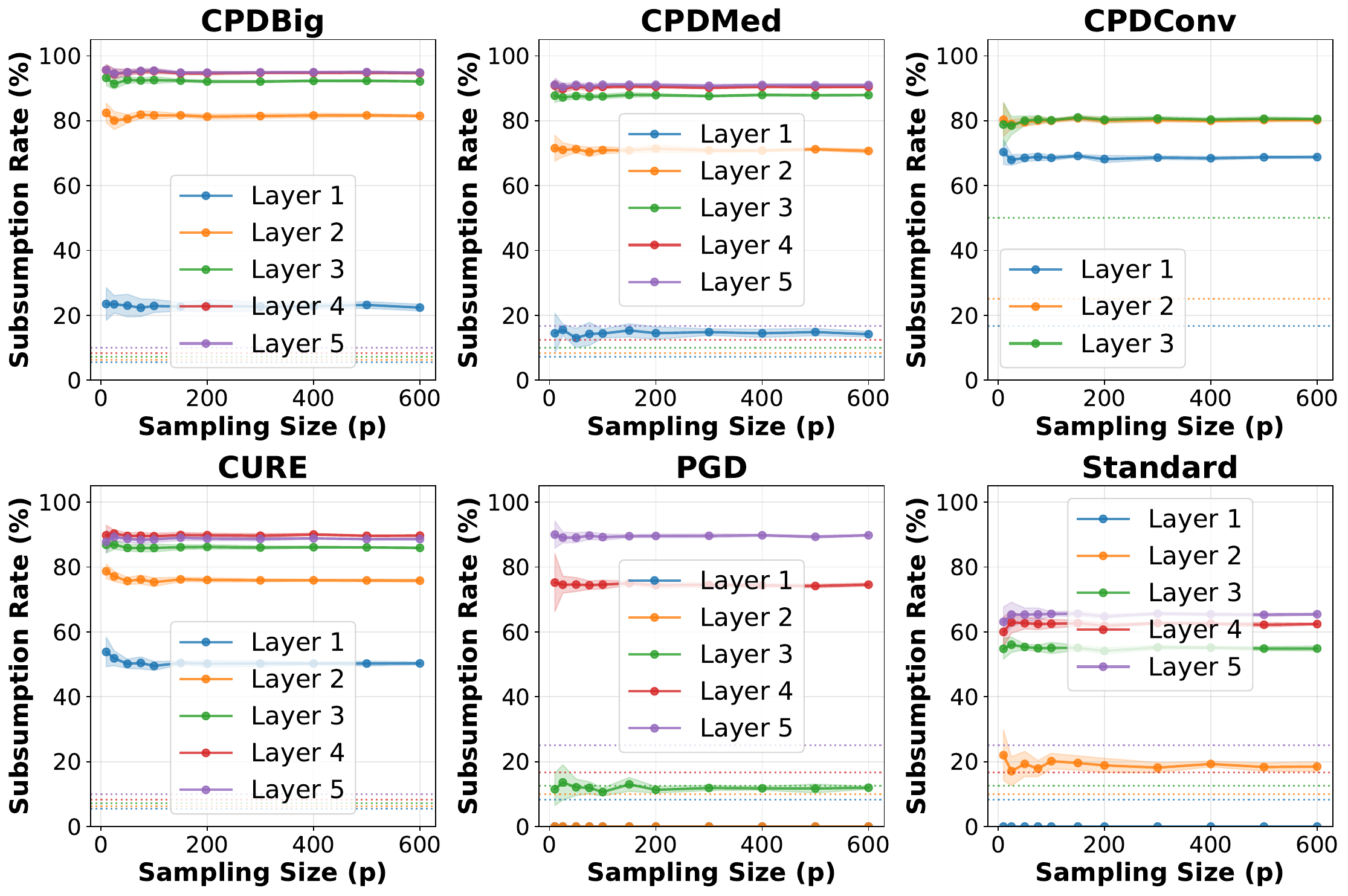}
    \caption{MNIST}
    \label{fig:mnist_sampling_1}
  \end{subfigure}
  
  \vspace{0.5em}
  
  \begin{subfigure}[b]{\textwidth}
    \centering
    \includegraphics[width=\textwidth]{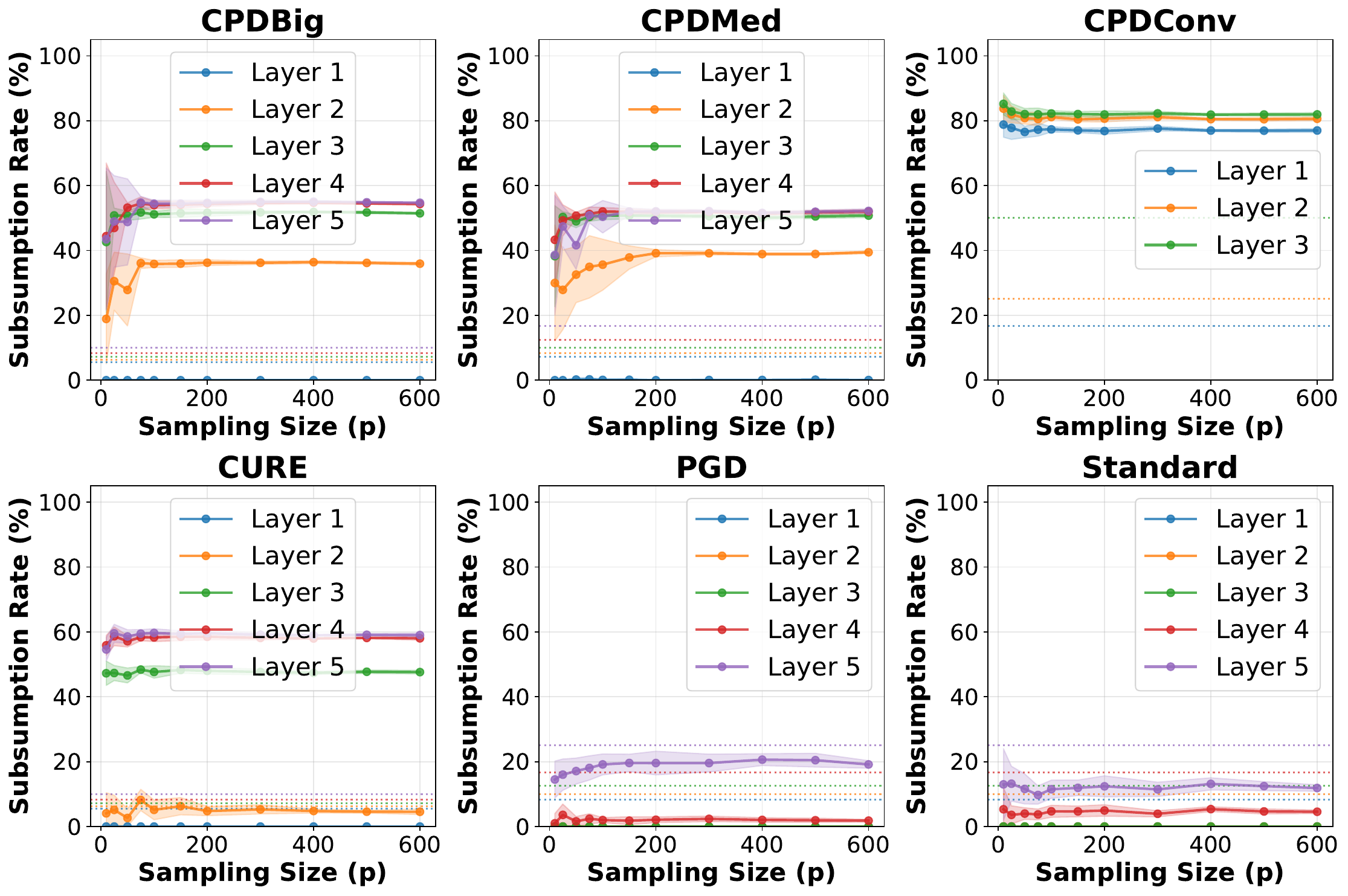}
    \caption{CIFAR-10}
    \label{fig:cifar_sampling_1}
  \end{subfigure}
  
  \caption{Comparing predicted minimum subsumption thresholds with observed subsumption rates across different sampling sizes with 1 template.}
  \label{fig:sampling_1_template}
\end{figure}

\begin{figure}[t!]
  \centering
  \begin{subfigure}[b]{\textwidth}
    \centering
    \includegraphics[width=\textwidth]{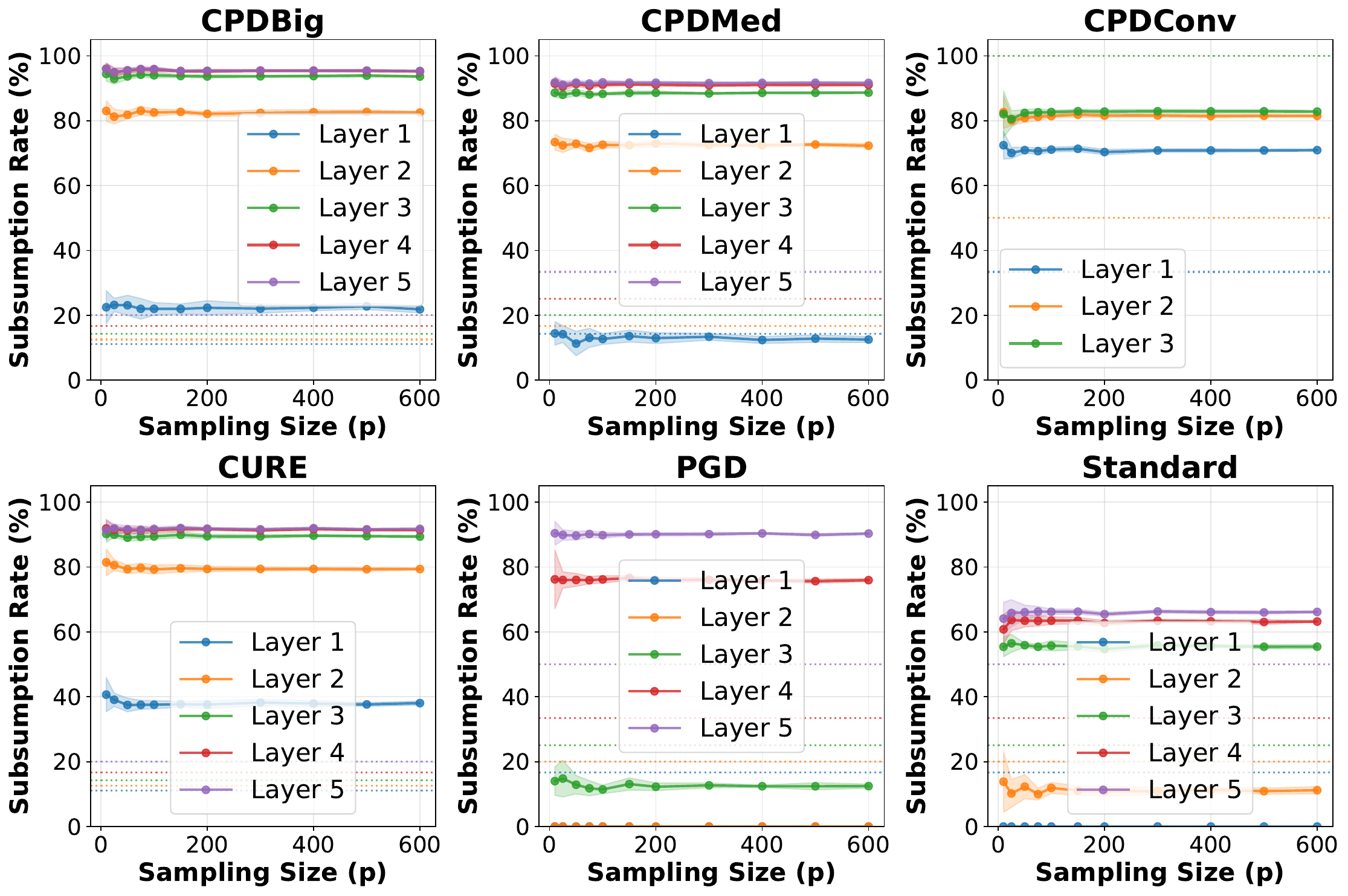}
    \caption{MNIST}
    \label{fig:mnist_sampling_2}
  \end{subfigure}
  
  \vspace{0.5em}
  
  \begin{subfigure}[b]{\textwidth}
    \centering
    \includegraphics[width=\textwidth]{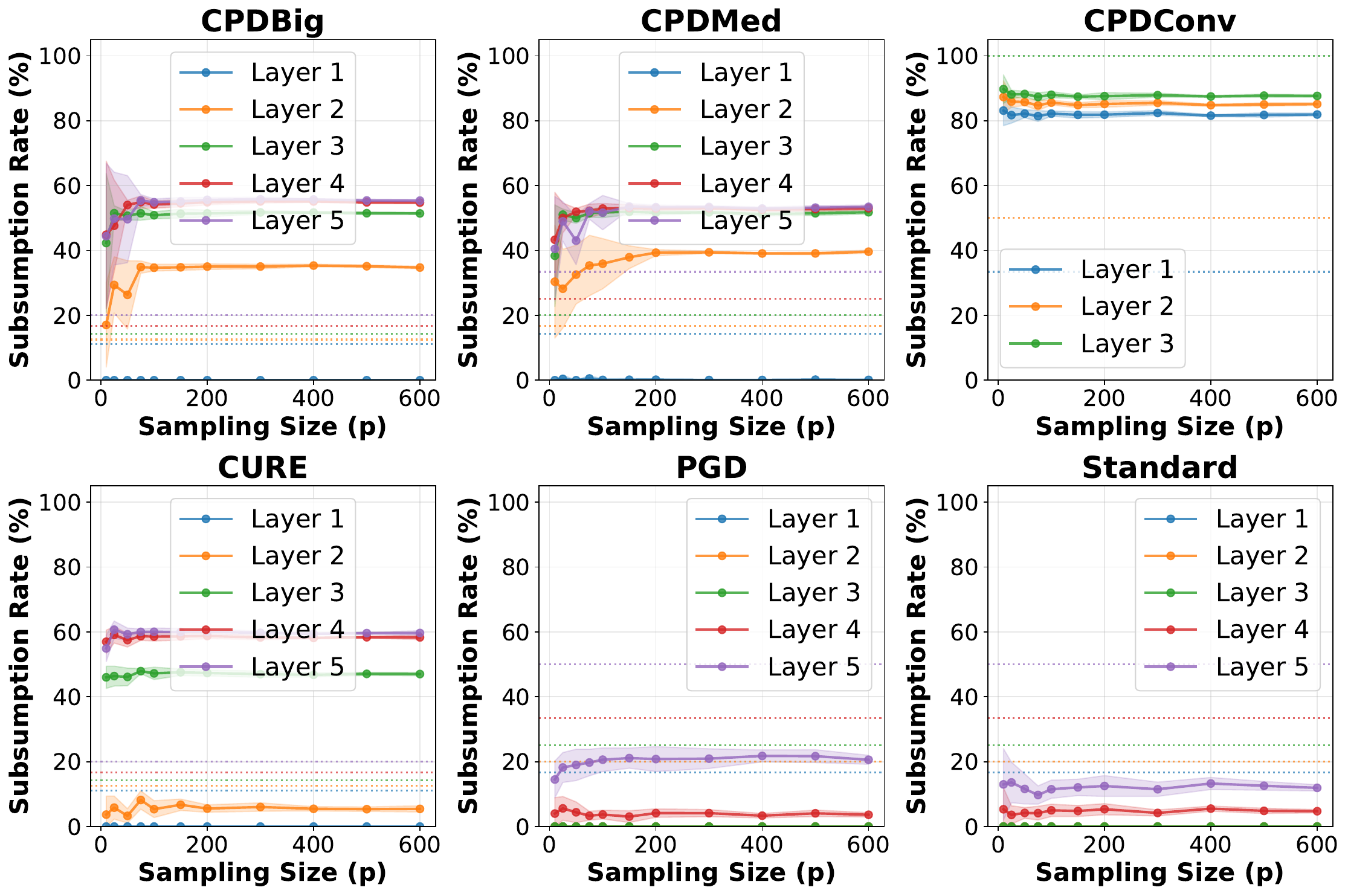}
    \caption{CIFAR-10}
    \label{fig:cifar_sampling_2}
  \end{subfigure}
  
  \caption{Comparing predicted minimum subsumption thresholds with observed subsumption rates across different sampling sizes with 2 templates.}
  \label{fig:sampling_2_templates}
\end{figure}

\begin{figure}[t!]
  \centering
  \begin{subfigure}[b]{\textwidth}
    \centering
    \includegraphics[width=\textwidth]{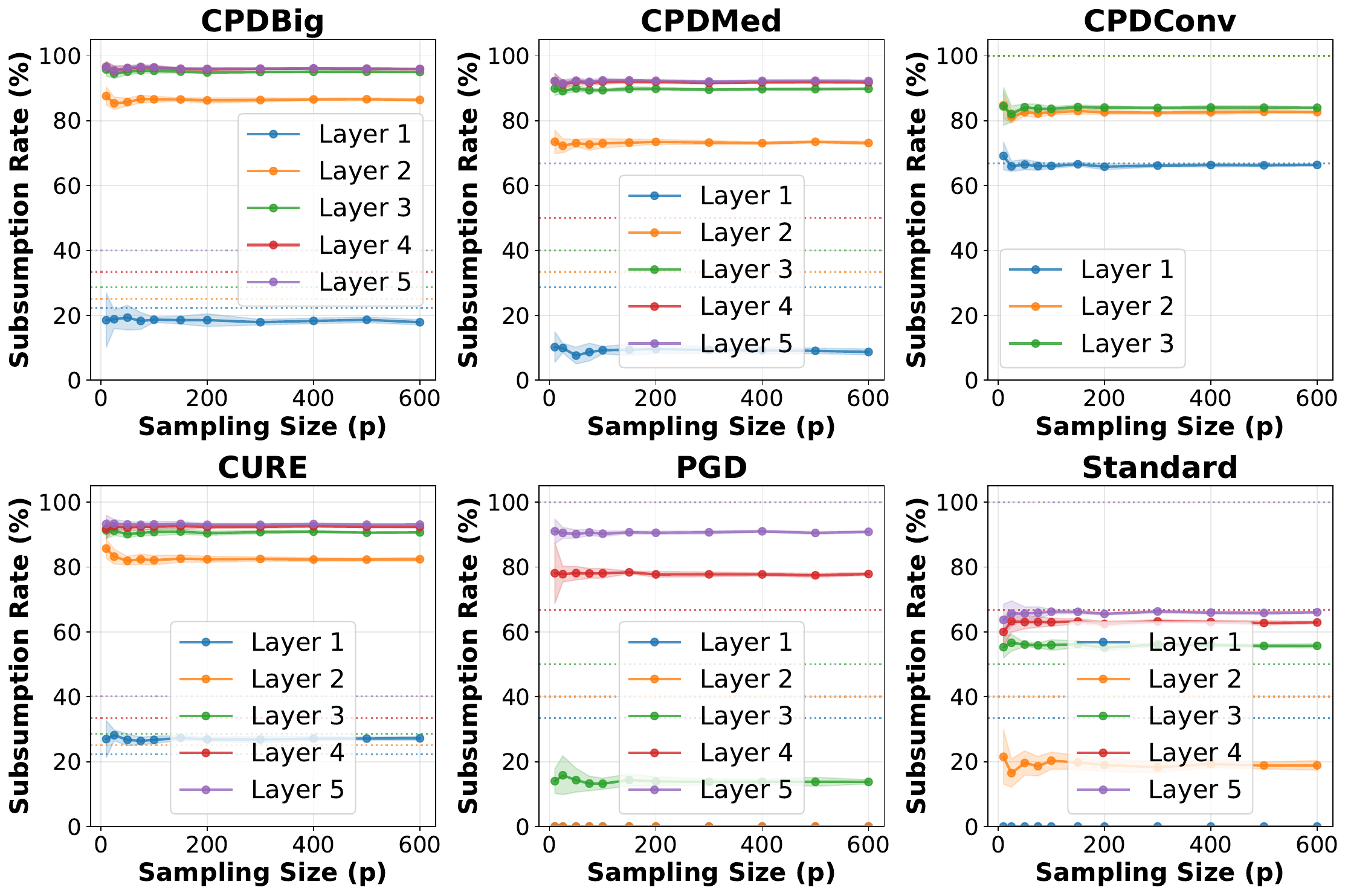}
    \caption{MNIST}
    \label{fig:mnist_sampling_4}
  \end{subfigure}
  
  \vspace{0.5em}
  
  \begin{subfigure}[b]{\textwidth}
    \centering
    \includegraphics[width=\textwidth]{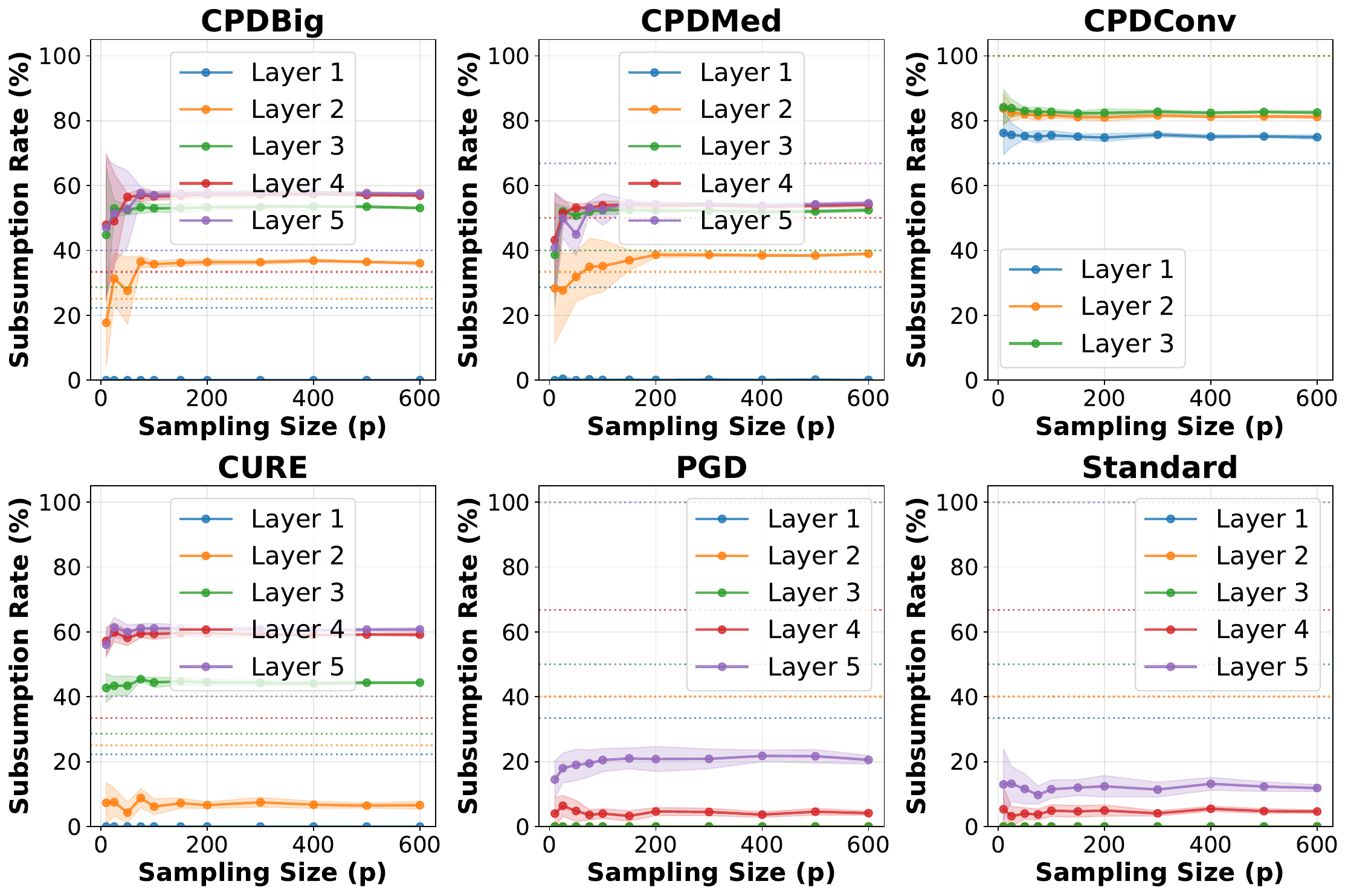}
    \caption{CIFAR-10}
    \label{fig:cifar_sampling_4}
  \end{subfigure}
  
  \caption{Comparing predicted minimum subsumption thresholds with observed subsumption rates across different sampling sizes with 4 templates.}
  \label{fig:sampling_4_templates}
\end{figure}

\Cref{fig:sampling_1_template,fig:sampling_2_templates,fig:sampling_4_templates}
fully detail our evaluation of how many sampled $L_0$ queries are needed to reliably estimate layer-wise subsumption for the algorithm of~\cref{sec:diagnostic}.
For each network, we select 10 input images (MNIST/CIFAR-10) and sample verification queries by perturbing 5 to 20
randomly chosen pixels. For each layer, we plot the estimated subsumption rate against a fixed set of templates as a function of
the number of sampled queries (10-600). For each configuration, we repeat this process 10 times and plot the mean (as solid lines) and standard deviation (as shaded regions).
We also plot the minimum subsumption rate required to yield speedup (dotted),
computed by \toolname's performance model (\cref{sec:speedup-model}). The plots are generated against different template counts of 1, 2, and 4.

Across networks and layers, subsumption estimates vary substantially for small samples (up to $\sim$150 queries),
but stabilize around $\sim$200 queries. Importantly, the threshold-crossing decision is not sensitive to sampling variability beyond this point (e.g.,~\cref{fig:pgd-mnist-p}).
While in rare cases, such as MNIST CPDMed with 2 templates~\cref{fig:mnist_sampling_2}, at layer 1, estimated subsumption falls very close to the threshold, even with larger samples of 600 queries, 
the variability remains similar. In almost all other cases, the gap between the estimated subsumption rate and the threshold to select or skip a layer for template reuse remains clear.

Thus we conclude that our choice of $P=200$ queries per image for subsumption estimation strikes a good balance between profiling overhead and reliable layer selection.

\end{document}